\RequirePackage{fix-cm}
\documentclass[twocolumn]{svjour3}          % twocolumn
\smartqed  % flush right qed marks, e.g. at end of proof
\usepackage{booktabs}
\usepackage{amsmath}
\usepackage{amssymb}
\usepackage{mathtools}
\usepackage{xspace}
\usepackage{hyperref}
\usepackage{graphicx}
\usepackage{caption}
\usepackage{subcaption}
\usepackage{multirow}
\usepackage{color, colortbl}
\usepackage{fancyvrb}
\usepackage{xcolor}
\usepackage{appendix}
\usepackage{natbib}

\newcommand{\eg}{\emph{e.g.,}\xspace}

\newcommand{\ie}{\emph{i.e.,}\xspace}

\newcommand{\reb}[1]{\textcolor{black}{#1}}
\newcommand{\oldreb}[1]{\textcolor{black}{#1}}

\definecolor{Gray}{gray}{0.9}
\definecolor{pink}{rgb}{1.0, 0.85, 0.85}

\journalname{International Journal of Computer Vision}

\hypersetup{pdfstartview={FitH}}
\hypersetup{pdfborder={0 0 0}}
\hypersetup{pdfpagemode={UseNone}}
\hypersetup{colorlinks=true}
\hypersetup{citecolor=blue}
\hypersetup{linkcolor=blue}

\begin{document}

%\title{Classifying and Localizing Unseen Actions from Objects}
%\title{Recognizing Unseen Actions from Object Awareness/Relations}
%\title{Spatial and Semantic Aware Object Embeddings for Unseen Action Recognition}

%CS: nog een suggestie
\title{Object Priors for Classifying and Localizing Unseen Actions}

%\title{Localizing Actions Without Examples through\\Spatial and Semantic Aware Object Embeddings%\thanks{Grants or other notes


\author{Pascal~Mettes \and William Thong \and
        Cees~G.~M.~Snoek %etc.
}


\institute{Pascal Mettes \at
Universiteit van Amsterdam, Amsterdam, the Netherlands\\
\email{P.S.M.Mettes@uva.nl}
\and
William Thong \at
Universiteit van Amsterdam, Amsterdam, the Netherlands\\
\email{W.E.Thong@uva.nl}
\and
Cees G. M. Snoek \at
Universiteit van Amsterdam, Amsterdam, the Netherlands\\
\email{cgmsnoek@uva.nl}
}

\date{Received: date / Accepted: date}

\maketitle

\begin{abstract}
This work strives for the classification and localization of human actions in videos, without the need for any labeled video training examples.
%Where existing work relies on global attribute or object scores and semantic matching to classify unseen actions, we also localize unseen actions in space and time.
\oldreb{Where existing work relies on transferring global attribute or object information from seen to unseen action videos, we seek to classify and spatio-temporally localize unseen actions in videos from image-based object information only.}
We propose three spatial object priors, which encode local person and object detectors along with their spatial relations. On top we introduce three semantic object priors, which extend semantic matching through word embeddings with three simple functions that tackle semantic ambiguity, object discrimination, and object naming. A video embedding combines the spatial and semantic object priors. It enables us to introduce a new video retrieval task that retrieves action tubes in video collections based on user-specified objects, spatial relations, and object size. Experimental evaluation on five action datasets shows the importance of spatial and semantic object priors for unseen actions. We find that persons and objects have preferred spatial relations that benefit unseen action localization, while using multiple languages and simple object filtering directly improves semantic matching, leading to state-of-the-art results for both unseen action classification and localization.
\end{abstract}
\section{Introduction}

%%% PAR 1: Establishing common ground.
The goal of this paper is to classify and localize human actions in video, such as \emph{shooting a bow}, \emph{doing a pull-up}, and \emph{cycling}. Human action recognition has a long tradition in computer vision, with initial success stemming from spatio-temporal interest points~\citep{chakraborty2012selective,laptev2005space}, dense trajectories \citep{wang2013dense,jain2013better}, and cuboids \citep{klaser2010human,liu2008recognizing}. Progress has recently been accelerated by deep learning, with the introduction of video networks exploiting two-streams~\citep{feichtenhofer2016convolutional,simonyan2014two} and 3D convolutions~\citep{carreira2017quo,tran2019video,zhao2018trajectory,feichtenhofer2019slowfast}. Building on such networks, current action localizers have shown the ability to detect actions precisely in both space and time, \eg~\citep{gkioxari2015finding,hou2017tube,kalogeiton2017action,zhao2019dance}. Common amongst action classification and localization approaches is the need for a substantial amount of annotated training videos. Obtaining training videos with spatio-temporal annotations~\citep{cheron2018flexible,mettes2019pointly} is expensive and error-prone, limiting the ability to generalize to any action. We aim for action classification and localization without the need for \emph{any} video examples during training.

%%% PAR 2: From language to zero-shot.
%In action recognition, many have explored the role of language,
\oldreb{In action recognition, many have explored the role of semantic action structures,}
from uncovering the grammar of an action~\citep{kuehne2014language} to enabling question answering in videos~\citep{zhu2017uncovering}. Language also plays a central role in zero-shot action recognition. Pioneering approaches transfer knowledge from attribute adjectives~\citep{liu2011recognizing,gan2016learning,zhang2015robust}, object nouns~\citep{jain2015objects2action}, or combinations thereof~\citep{wu2014zero}. The supervised action recognition literature has already revealed the strong link between actions and objects for recognition~\citep{gupta2007objects,jain201515,wu2007scalable}. Especially when object classification scores are obtained from large-scale image datasets~\citep{deng2009imagenet,lin2014microsoft} and matched with any action through word embeddings~\citep{grave2018learning}. We follow this object-based perspective for unseen actions. We add a generalization to spatio-temporal localization, by including local object detection scores and prior knowledge about prepositions, and we examine the linguistic relations between actions and objects to improve their semantic matching.

%%%  PAR 3: Spatial-aware embedding.
Our first contribution are three spatial object priors that encode local object and actor detections, as well as their spatial relations. We are inspired by the supervised action classification literature, where the spatial link with objects is well established, \eg~\citep{gupta2007objects,kalogeiton2017joint,moore1999exploiting,wu2007scalable,yao2011classifying}. To incorporate information about spatial prepositions without action video examples, we start from existing object detection image datasets and models. Box annotations in object datasets allow us to assess how people and objects are commonly related spatially. From discovered spatial relations, we propose a score function that combines person detections, object detections, and their spatial match for unseen action classification and localization. The spatial priors were previously introduced in the conference version~\citep{mettes2017spatial} preceding this paper.

%%% PAR 4: Semantic-aware embedding.
Our second contribution, not addressed in~\citep{mettes2017spatial}, are three semantic object priors. Common in unseen action recognition using objects is to estimate relations using word embeddings~\citep{chang2016dynamic,jain2015objects2action,li2019zero,wu2016cvpr}. They provide dense representations on which similarity functions are performed to estimate semantic relations~\citep{mikolov2013distributed}. Similarities from word embeddings have several linguistic limitations relevant for unseen actions. Our semantic priors address three limitations with simple functions on top of word embedding similarities. First, we leverage word embeddings across languages to reduce semantic ambiguity in the action-object matching. Second, we show how to filter out non-discriminative objects directly from similarities between all objects and actions. Third, we show how to focus on basic-level names in object datasets to improve relevant matching. We combine the spatial and semantic object priors into a video embedding.

%\cs{Need one closing sentence on technical combination of all six. This also needs to be added to abstract. Would this still be called an embedding? An object prior representation, something else?}

%%% PAR 5: Empirical findings and new task.
Experiments on five action datasets demonstrates the effectiveness of our six object priors. We find that the use of prepositions in our spatial-aware embedding enables effective unseen action localization using only a few localized objects. Our semantic object priors improve both unseen action classification and localization, with multi-lingual word embeddings, object discrimination functions, and a bias towards basic-level objects for selection. We also introduce a new task, action tube retrieval, where users can search for action tubes by specifying desired objects, sizes, and prepositions. Our object prior embedding obtains state-of-the-art zero-shot results for both unseen action classification and localization, highlighting its effectiveness and more generally, emphasizing the strong link between actions and objects.

%%% PAR 6: Paper organization.
The rest of the paper is organized as follows. Section~\ref{sec:related} discusses related work. Sections \ref{sec:method-localization} and \ref{sec:method-classification} detail our spatial and semantic object priors. Sections~\ref{sec:setup} and~\ref{sec:results} discuss the experimental setup and results. The paper is concluded in Section~\ref{sec:conclusions}.

\section{Related work}
\label{sec:related}

%\cs{Where is the praise?}

%\cs{Zijn alle refs uit Table 6, 7 en 8 gecovered?} \psmm{All except supervised ones form UCF Sports table now discussed.}

\subsection{Unseen action classification}
For unseen action classification, a common approach is to generalize from seen to unseen actions by mapping videos to a shared attribute space~\citep{gan2016learning,liu2011recognizing,zhang2015robust}, akin to attribute-based approaches in images~\citep{lampert2013attribute}. Attribute classifiers are trained on seen actions and applied to test videos. The obtained attribute classifications are in turn compared to a priori defined attribute annotations. With the use of attributes, actions not seen during training can still be recognized. The attribute-based approach has been extended by using knowledge about test video distributions in transductive settings~\citep{fu2015transductive,xu2017transductive} and by incorporating domain adaptation~\citep{kodirov2015unsupervised,xu2016multi}. While enabling zero-shot recognition, attributes require prior expert knowledge for every action, which does not generalize to arbitrary queries. 
%Moreover, attributes are defined globally, making localization infeasible.
%Since attributes are a priori defined from expert knowledge, this setting does not generalize to arbitrary zero-shot queries. Furthermore, the attributes are globally defined, making localization impossible.
Hence we refrain from employing attributes.

Several works have investigated skipping the intermediate mapping to attributes by directly mapping unseen actions to seen actions. \cite{li2016recognizing} and \cite{tian2018zero} map features from videos to a semantic space shared by seen and unseen actions, while \cite{gan2016recognizing} train a classifier for unseen actions by performing several levels of relatedness to seen actions. Other works propose to synthesize features for unseen actions~\citep{mishra2018generative,mishra2020zero}, learn a universal representation of actions~\citep{zhu2018towards}, or differentiate seen from unseen actions through out-of-distribution detection~\citep{mandal2019out}. All these works eliminate the need for attributes for unseen action classification. We also do not require attributes for our action classification, yet with the same model, we also enable action localization.

Several works have considered object classification scores for their zero-shot action, or event, classification by performing a semantic matching through word vectors~\citep{an2019spatiotemporal,bishay2019tarn,chang2016dynamic,inoue2016adaptation,li2019zero,jain2015objects2action,wu2016cvpr} or auxiliary textual descriptions~\citep{gan2016concepts,habibian2016video2vec}. Objects provide an effective common space for unseen actions, as object scores are easily obtained by pre-training on existing large-scale datasets, such as ImageNet~\citep{deng2009imagenet}. Objects furthermore allow for a generalization to arbitrary unseen actions, since relevant objects for new actions can be obtained on-the-fly through word embedding matching with object names. In this work, we follow this line of work and generalize to spatio-temporal localization by modeling the spatial relations between actors and objects. This allows us to perform action classification and localization within the same approach. \oldreb{Different from the common setup for zero-shot actions~\citep{junior2019zero}, we do not assume access to any training videos of seen actions. We seek to recognize actions in video without ever having seen a video before, solely by relying on prior knowledge about objects in images and their relation to actions.}

%\cs{Deze paragraaf voelt een beetje als mosterd na de maaltijd, als het zo belangrijk is waarom dan maar zo weinig refs? Ik zou dit anders aanvliegen.} Due to the importance of language in unseen action recognition, 
%
To improve semantic matching, \cite{alexiou2016exploring} correct class names to increase unseen action discrimination. Similar in spirit are approaches that employ query expansion~\citep{dalton2013zero,de2016knowledge} or textual action descriptions \citep{gan2016recognizing,habibian2016video2vec,wang2017alternative} to make the action inputs more expressive. In contrast, we focus on improving the semantic matching itself to deal with semantic ambiguity, non-discriminative objects, and object naming.

%\psmm{other related papers and our differences.}
%\psmm{first paper.} \psmm{other papers on this topic.} \psmm{relation to this paper.}

%
%\cs{Losse flodder} 
%\psmm{De vraag is of we los nog een paper als deze, die iets doet met semantics, moeten noemen hier. Je kan het punt maken dat het verschil met huidige werken op dit punt al duidelijk genoeg is. Voor nu weggelaten.}Rather than focusing on improving class names themselves to increase unseen action discrimination~\citep{alexiou2016exploring}, we improve the semantic matching between actions and objects. \cs{Dat lijkt me wel, ik zou liever meer dan 1 paper zien. Misschien is dat meer iets voor intro? Het staat hier nu een beetje verloren.}

%\cs{I miss a paper by Shah on clustering without examples.}

%\cs{Maybe add a section on weakly-supervised localization? Some are mentioned in Table 7. Or work that localizes actions with the aid of language? Video Q\&A?}

\subsection{Unseen action localization}
Spatio-temporal localization of actions without examples is hardly investigated in the current literature. \cite{jain2015objects2action} split each test video into spatio-temporal proposals~\citep{jain2017tubelets}. Then for each proposal, boxes are sampled and individually fed to a pre-trained object classification network to obtain object scores. The object scores of each proposal are semantically matched to the action and the best matched proposal is selected as the location of interest. In this paper, we employ local object detectors and embed spatial relations between humans and objects. Where~\cite{jain2015objects2action} implicitly assume that the spatial location of objects and the humans performing actions is identical, our spatial object priors explicitly model how humans and objects are spatially related, whether objects are above, to the left, or on the human. Moreover, we go beyond standard word embedding similarities for semantic matching between actions and objects to improve both unseen action classification and localization. \cite{soomro2017unsupervised} investigate action localization in an unsupervised setting, which discriminatively clusters similar action tubes but does not specify action labels. In contrast, we seek to discover both action locations and action labels without training examples or manual action annotations.

\oldreb{Several works have investigated unseen action localization in the temporal domain. \citep{zhang2020zstad} perform zero-shot temporal action localization by transferring knowledge from temporally annotated seen actions to unseen actions. \cite{jain2020actionbytes} learn an action localization model from seen actions in trimmed videos, enabling zero-shot temporal action localization by a semantic knowledge transfer of unseen actions. \cite{sener2018unsupervised} learn to temporally segment actions in long videos in an unsupervised manner. Different from these works, we perform unseen action localization in space and time simultaneously.}
%\cs{Ik snap het contrast met  \cite{soomro2017unsupervised} niet}
%\psmm{Kalogeiton 2017 joint actor and object localization can also be here.}
%\psmm{To do: also incorporate contribution 2. Likely by rewriting previous sentence too.}
%\cs{What about your second contribution, the semantic matching? You don't say anything about it.}

%\subsection{Action-object relations}
%\cs{The division in related work does not make much sense to me, why is this topic here? Isn't it better to have the sentence: ``Action representation \ldots annotations'' in the paragraph with the first contribution?}
%
%We are inspired by supervised action classification literature, where the spatial link with objects is well established, see \eq~\citep{gupta2007objects,mettes2017localizing,moore1999exploiting,wu2007scalable,yao2011classifying}. Action representations and spatial relations with objects can be jointly modelled, for example with local features~\citep{escorcia2013spatio} or in a deep network~\citep{zhang2019structured}. In this manner, these works are capable of performing action recognition in supervised settings. Here, we avoid the need for labelled action videos an/or human-object annotations. We propose object embeddings that exploit spatial and semantic relations with objects to localize, classify, and retrieve unseen actions, by building on top of freely available object and actor detectors.

\reb{
\subsection{Self-supervised video learning}
Recently, a number of works have proposed approaches for representation learning for unlabeled videos through self-supervision. The general pipeline is to train a pre-text task on unlabeled data and transfer the knowledge to a supervised downstream task~\citep{jing2020self} or by clustering video datasets without manual supervision~\citep{asano2020labelling}. Pretext tasks include dense predictive coding~\citep{han2020memory}, shuffling frames~\citep{fernando2017self,xu2019self}, exploiting spatial and/or temporal order~\citep{jenni2020video,tschannen2020self,wang2019self}, or by matching frames with other modalities~\citep{afouras2020self,alayrac2020self,owens2018audio,patrick2020multi}. Self-supervised approaches utilize unlabeled train videos to learn representations without semantic class labels. In contrast, we do not use any training videos and instead classify and localize actions using object classes and bounding boxes from images. Since we do not assume any video knowledge, common losses and notions from the zero-shot and self-supervised literature can not be leveraged. It is  the  object  priors  that  still  allow  us  to  classify and spatio-temporally localize unseen actions in videos.
}
\section{Spatial object priors}
\label{sec:method-localization}

In unseen action localization, the aim is to discover a set of spatio-temporal action tubes from test videos for each action in the set of all actions $\mathcal{A} = \{A_1,\dots,A_C\}$, with $C$ the total number of actions. Furthermore, unseen action classification is concerned with predicting the label of each test video from $\mathcal{A}$. For each action, nothing is known except its name. The evaluation is performed on a set of $N$ unlabeled and unseen test videos denoted as $\mathcal{V}$. In this section, we outline how to obtain such a localization and classification with spatial priors from local objects using prior knowledge. % about prepositions.
%In the next section, we show how to extend the embedding by incorporating global contextual objects with semantic awareness.

\subsection{Priors from persons, objects, and prepositions}
For a test video $v \in \mathcal{V}$ and unseen action $a \in \mathcal{A}$, the first step of our approach is to score local boxes in the video with respect to $a$. For a bounding box $b$ in video frame $F$, we define a score function $s(\cdot)$ for action class $a$. The score function is proportional to three priors.
\\\\
\textbf{Object prior I \oldreb{(person prior)}:} \emph{The likelihood of any action in} b \emph{is proportional to the likelihood of a person present in} b.
\\\\
The first prior follows directly from our human action recognition task. The first condition is independent of the specific action class, as it must hold for any action. The score function therefore adheres to the following:
\begin{equation}
s(b, F, a) \propto Pr(\texttt{person} | b). 
\end{equation}
\\
\textbf{Object prior II \oldreb{(object location prior)}:} \emph{The likelihood of action} $a$ {in box} b \emph{is proportional to the likelihood of detected objects that are (i) semantically close to action class $a$ and (ii) the detection is sufficiently close to} b.
\\\\
The second prior states that the presence of an action in a box $b$ also depends on the presence of relevant objects in the \emph{vicinity} of $b$. We formalize this as:
\begin{equation}
s(b, F, a) \propto \sum_{o \in \mathcal{L}} \Psi(o, a) \cdot \max_{b' \in o_{D}(F,b)} Pr(o | b'),
\label{eq:emb-c2}
\end{equation}
where $\mathcal{L}$ denotes the set of pre-trained object detections and $o_{D}(F,b)$ denotes the set of all object detections of object $o$ in frame $F$ that are near to box $b$.
%In practice, this includes all boxes for which edges are within 25 pixels of $b$, although the localization and classification scores are hardly affected by other settings, as long as the value is smaller than the whole frame.
\oldreb{Empirically, the second object prior is robust to the pixel distance to determine the neighbourhood set $o_D(F,b)$ for box $b$, as long as it is a non-negative number smaller than the frame size. We use a value of 25 throughout.}
Function $\Psi(o,a)$ denotes the semantic similarity between object $o$ and $a$ and is defined as the word embedding similarity:
\begin{equation}
\Psi(o,a) = \cos(\phi(o), \phi(a)),
\end{equation}
with $\phi(\cdot) \in \mathbb{R}^{300}$ the word embedding representation. The word embeddings are given by a pre-trained word embedding model, such as word2vec~\citep{mikolov2013distributed}, FastText~\citep{grave2018learning}, or GloVe~\citep{pennington2014glove}.
\\\\
\textbf{Object prior III \oldreb{(spatial relation prior)}:} \emph{The likelihood of action $a$ in} b \emph{given an object} o \emph{with box detection} d \emph{that abides object prior II, is proportional to the match between the spatial awareness of} b \emph{and} d \emph{with the prior spatial awareness of $a$ and} o.
\\\\
The third prior incorporates spatial awareness between actions and objects. We exploit the observation that people interact with objects in preferred spatial relations. We do this by gathering statistics from the same image dataset used to pre-train the object detectors. By reusing the same dataset, we keep the amount of knowledge sources contained to a dataset for object detectors and a semantic word embedding. For the spatial relations, we examine the bounding box annotations for the person class and all object classes. We gather all instances where an object and person box annotation co-occur. We quantize the gathered instances into representations that describe coarse spatial prepositions between people and objects.

\begin{figure}[t]
\centering
\includegraphics[width=\linewidth]{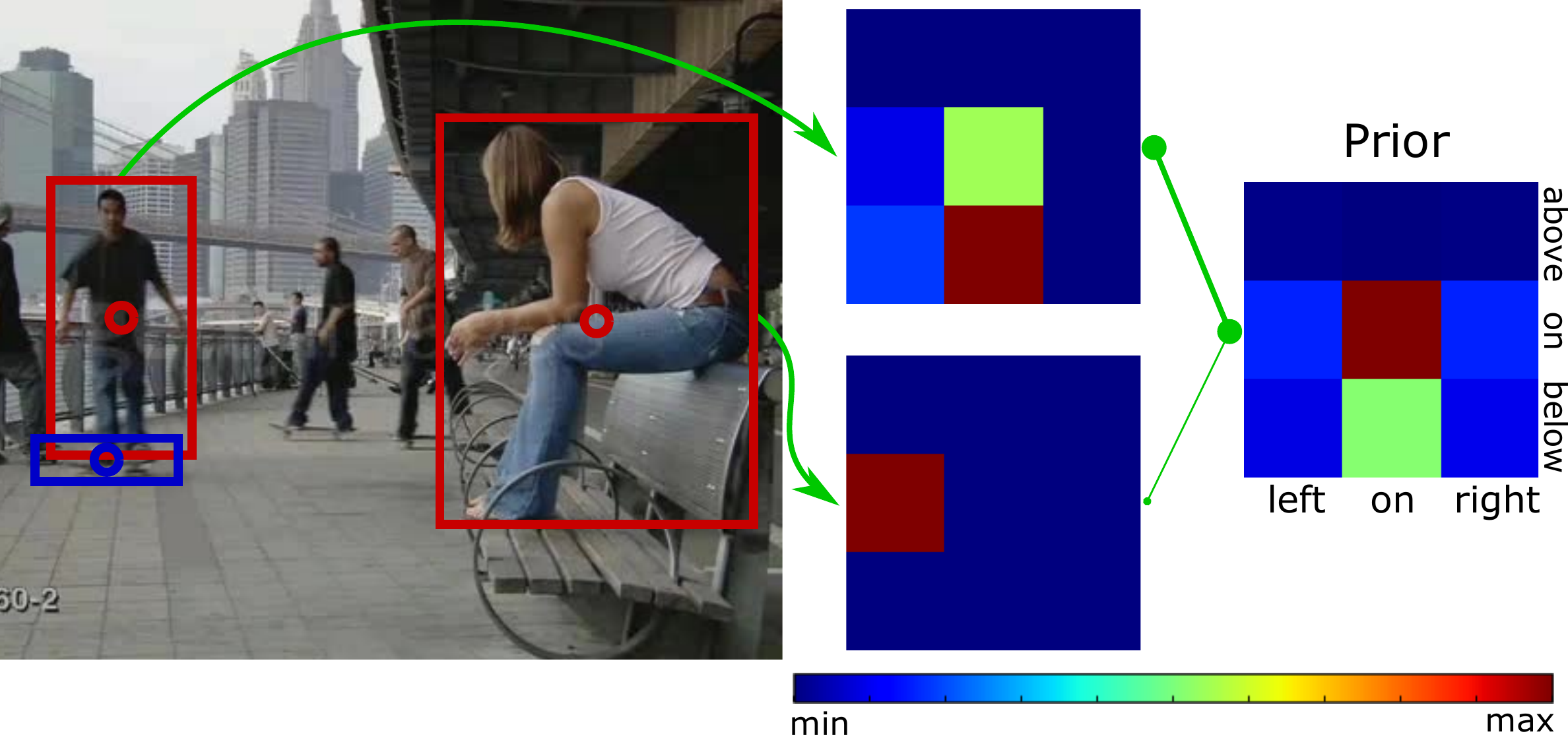}
\caption{\oldreb{Intuition behind spatial object priors. The spatial relations (end of green arrows) of the two persons (red boxes) have different spatial relations with the detected skateboard (blue box). The spatial relations for the person on the left are a better match with the spatial relations obtained from prior knowledge. This match enforces the likelihood that the person on the left is involved in a skateboarding activity.}}
%Intuition behind spatial object priors. The two persons (red) have different spatial relations with the detected skateboard (blue). 
%The spatial relations for the person on the left are a better match with the spatial relations obtained from prior knowledge. This match enforces the likelihood that the person on the left is involved in s skateboarding activity.}
%aggregated relations from prior knowledge, enforcing the likelihood of actions involving a skateboard for that person.}
\label{fig:spatial-intuition}
\end{figure}

The spatial relation between an object box relative to a person box is quantized into a 9-dimensional grid. This grid represents how the object box is spatially distributed to the person box with respect to the following prepositions: $\{$\emph{above left}, \emph{above}, \emph{above right}, \emph{left}, \emph{on}, \emph{right}, \emph{below left}, \emph{below}, \emph{below right}$\}$. Since no video examples are given in our setting, prepositions can only be obtained from prior image sources and we therefore exclude relations such as \emph{in front of} and \emph{behind of}.
%
%\cs{Hier nog kort iets zeggen over 2D beperking, en dat we geen 3D relaties kunnen? In front of/behind.}
%
Let $d_{1}(b, d) \in \mathcal{R}^{9}$ denote the spatial distribution of object box $d$ relative to person box $b$. Furthermore, let $d_{2}(\texttt{person}, o)$ denote the gathered distribution of object $o$ with respect to a person from the image dataset. We define the spatial relation function as:
\begin{equation}
\Phi(b, d, o) = 1 - \text{JSD}_{2}(d_{1}(b, d) || d_{2}(\texttt{person}, o)),
\label{eq:jsd}
\end{equation}
where $\text{JSD}_{2}(\cdot||\cdot) \in [0,1]$ denotes the Jensen-Shannon Divergence with base 2 logarithm. Intuitively, this function determines the extent to which the 9-dimensional distributions match, as visualized in Figure~\ref{fig:spatial-intuition}. The more similar the distributions, the lower the divergence, and the higher the score according to Equation~\ref{eq:jsd}.
\\\\
\textbf{Combined spatial priors.} Our final box score combines the priors of persons, objects, and spatial prepositions. We combine the three priors into the following score function for a box $b$ with respect to action $a$:
\begin{equation}
\begin{split}
s(b, F, a) = & Pr(\texttt{person} | b) + \sum_{o \in \mathcal{O}} \Psi(o, a) \cdot \\
& \max_{b' \in o_{D}(F, b)} \bigg( Pr(o | b') \cdot \Phi(b, b', o) \bigg).
\end{split}
\label{eq:embedding}
\end{equation}

%%%%%%%%%%%%%%%%%%%%%%%%%%%%%%%%%%%%%%%%%%%%%%%%%%
% LINKING
%%%%%%%%%%%%%%%%%%%%%%%%%%%%%%%%%%%%%%%%%%%%%%%%%%
\subsection{Linking action tubes}
\label{sec:m-linking}
Given scored boxes in individual frames, we link boxes into tubes to arrive at a spatio-temporal action localization. We link boxes that have high scores from our object embeddings and have a high spatial overlap. Given an action $a$ and boxes $b_{1}$ and $b_{2}$ in consecutive frames $F_{1}$ and $F_{2}$, the link score is given as:
\begin{equation}
w(b_{1}, b_{2}, a) = s(b_{1}, F_{1}, a) + s(b_{2}, F_{2}, a) + \text{iou}(b_{1}, b_{2}),
\end{equation}
where $\text{iou}(\cdot, \cdot)$ states the spatial intersection-over-union score. We solve the problem of linking boxes into tubes with the Viterbi algorithm~\citep{gkioxari2015finding}.
\oldreb{For a video $V$, we apply the Viterbi algorithm on the link scores to obtain spatio-temporal action tubes. In each tube, we continue linking as long as there is at least one box in the next frame with an overlap higher than 0.1 and with a combined action score of at least 1.0. Otherwise we stop linking. Incorporating the stopping criterion allows us to localize actions in time also, akin to~\citep{gkioxari2015finding}.} We reiterate this process until we obtain $T$ tubes. The action score for $a$ of an action tube $t$ is defined as the average score of the boxes in the tube:
\begin{equation}
\ell_\text{tube}(t, a) = \frac{1}{|t|} \sum_{i=1}^{|t|} s(b_{t_i}, F_{t_i}, a),
\label{eq:tubescore}
\end{equation}
where $b_{t_i}$ and $F_{t_i}$ denote respectively the box and frame of the $i^{\text{th}}$ element in $t$.
\\\\
\textbf{Unseen action localization and classification.}
For unseen action localization, we gather tubes across all test videos and rank the tubes using the scores provided by Equation~\ref{eq:tubescore}. We can also perform unseen action classification using the spatial priors by simply disregarding the tube locations. For each video, we predict the action class label as the action with the highest tube score within the video.
\subsection{Action tube retrieval}
\label{sec:method-retrieval}
The use of objects with spatial priors extends beyond unseen action classification and localization. We can also perform a new task, dubbed action tube retrieval. This task resembles localization, as the goal is to rank the most relevant tubes the highest. Different from localization, we now have the opportunity to specify which objects are of interest and which spatial relations are desirable for a detailed result. Furthermore, inspired by the effectiveness of size in actor-object relations~\citep{escorcia2013spatio}, we extend the retrieval setting by allowing users to specify a desired relative size between actors and objects. The ability to specify the object, spatial relations, and size allows for different localizations of the same action. To enable such a retrieval, we extend the box score function of Equation~\ref{eq:embedding} as follows:
\begin{equation}
\begin{split}
s(b, & F, o, r, s) = Pr(\texttt{person} | b) + \max_{b' \in o_{D}(F, b)}\\
& \bigg( Pr(o | b') \cdot \Phi_r(b, b', r) \cdot \big( 1 - |\frac{\text{size}(b')}{\text{size}(b)} - s| \big) \bigg),
\end{split}
\label{eq:embedding-retrieval}
\end{equation}
where $o$ denotes the user-specified object, $r \in \mathbb{R}^9$ the specified spatial relations, and $s$ the specified relative size. The spatial relation function is modified to directly match box relations to specified relations:
\begin{equation}
\Phi_r(b, d, r) = 1 - \text{JSD}_{2}(d_{1}(b, d) || r).
\end{equation}
With the three user-specified objectives, we again score individual boxes first and link them over time. The tube score is used to rank the tubes across a video collection to obtain the final retrieval result.
%%%%%%%%%%%%%%%%%%%%%%%%%%%%%%%%%%%%%%%%%%%%%%%%%%
% CLASSIFICATION
%%%%%%%%%%%%%%%%%%%%%%%%%%%%%%%%%%%%%%%%%%%%%%%%%%
\section{Semantic object priors}
\label{sec:method-classification}
\begin{figure*}[t]
\centering
\begin{subfigure}{0.30\textwidth}
\includegraphics[width=\textwidth]{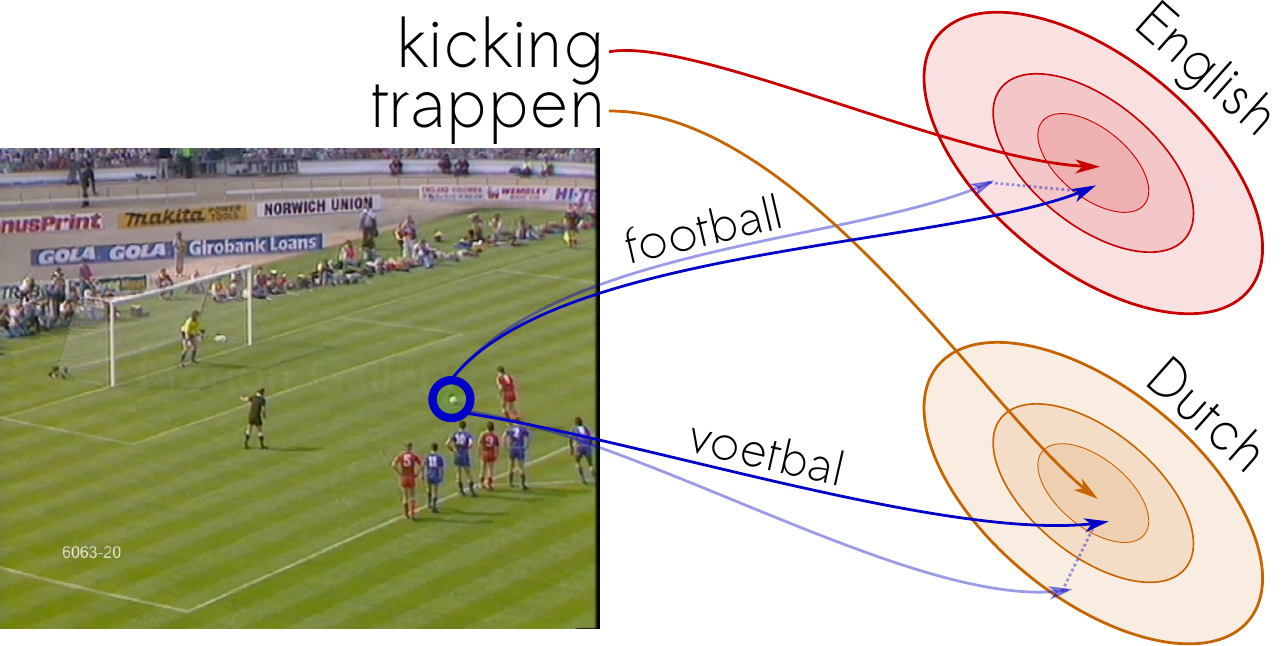}
\caption{\textit{Semantic disambiguation.}}
\end{subfigure}
\hspace{0.5cm}
\begin{subfigure}{0.30\textwidth}
\includegraphics[width=\textwidth]{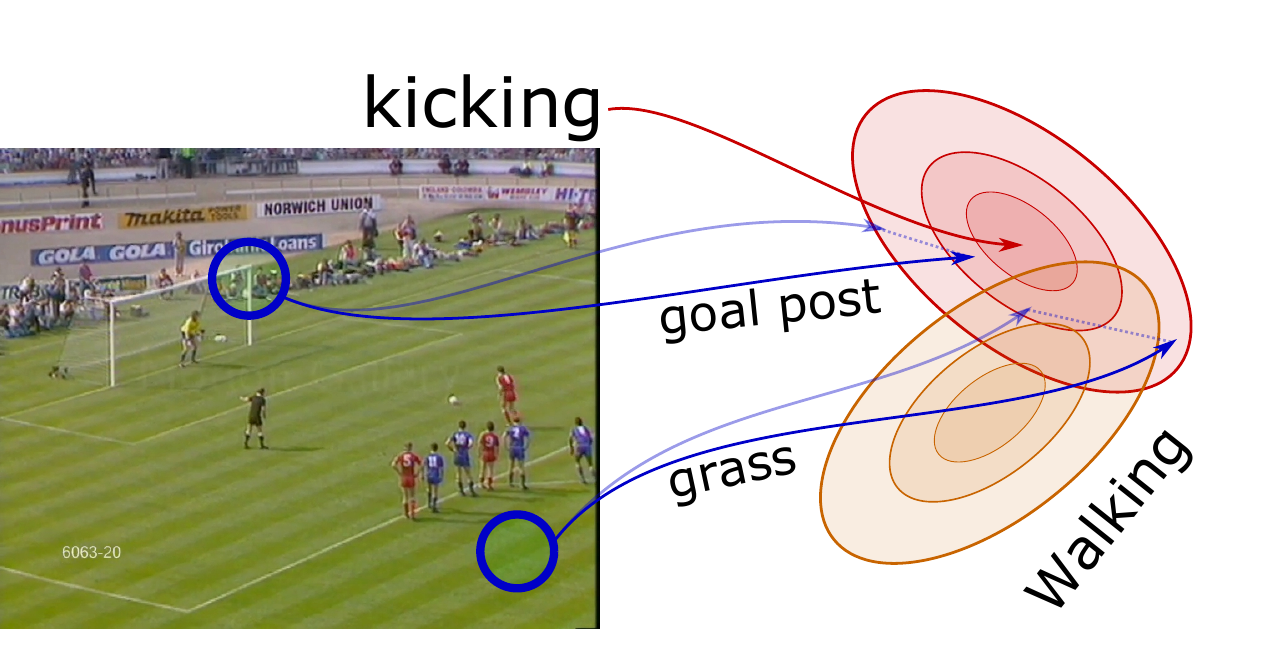}
\caption{\textit{Object discrimination.}}
\end{subfigure}
\hspace{0.5cm}
\begin{subfigure}{0.30\textwidth}
\includegraphics[width=\textwidth]{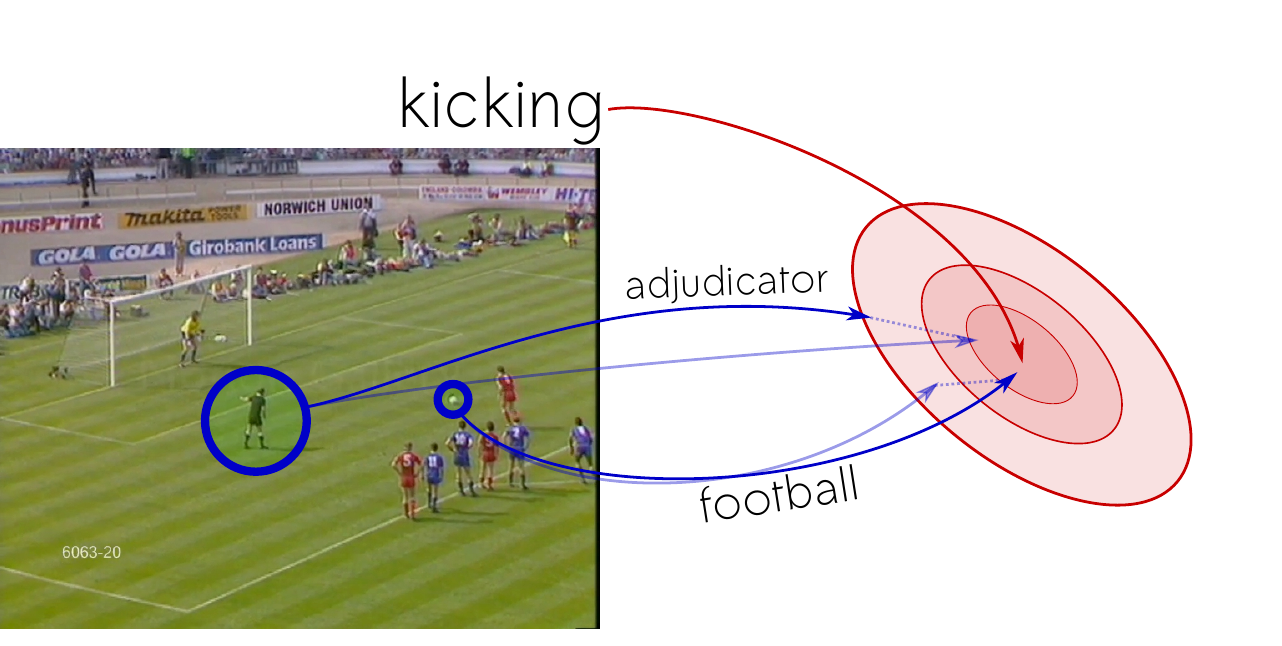}
\caption{\textit{Object naming.}}
\end{subfigure}
\caption{Intuition behind our three semantic object priors.
%In all cases, the closer to the center of the action embedding (in red and orange), the higher the object relevance. 
\oldreb{The red and orange distributions denote the word embeddings of the action kicking in English and Dutch. The closer to the center an object is, the higher the semantic similarity to the action.}
In (a), the object football is enforced, because its semantic similarity is high across languages, reducing semantic ambiguity. In (b), the importance of grass is decreased, as it is also relevant for another action, while the opposite happens for goal post. In (c), the importance of football is increased and of adjudicator decreased, as football follows basic-level object naming, in contrast to adjudicator (a referee).}
\label{fig:illustration-semantic}
\end{figure*}

Spatial object priors relying on local objects enables a spatio-temporal localization of unseen actions. However, local objects do not tell the whole story. When a person performs an action, this is typically happens in a suitable context. Think about someone \emph{playing tennis}. While the tennis racket provides a relevant cue about the action and its location, surrounding objects from context, such as tennis court and tennis net, further enforce the action likelihood. Here, we add three additional object priors to integrate knowledge from global objects for unseen action classification and localization. We start from the common word embedding setup for semantic matching, which we extend with three simple priors that make for effective unseen action matching with global objects. Lastly, we outline how to integrate the semantic and spatial object priors for unseen actions. Figure~\ref{fig:illustration-semantic} illustrates our proposal.

\subsection{Matching and scoring with word embeddings}
To obtain action scores for a video $v \in \mathcal{V}$, the common setup is to directly use the object likelihoods from a set of global objects $\mathcal{G}$ and their semantic similarity. Since $\mathcal{G}$ typically contains many objects, the usage is restricted to the objects with the highest semantic similarity to action $a$:
\begin{equation}
\Psi(g, a) = \cos(\phi(g), \phi(a))~\text{such that}~g \in \mathcal{G}_a,
\label{eq:semsim}
\end{equation}
where $\mathcal{G}_a$ the set of $k$ most similar objects with respect to $a$. The video score function is defined as:
\begin{equation}
\ell_\text{video}(v, a) = \sum_{g \in \mathcal{G}_a} \Psi(g, a) \cdot Pr(g|v),
\end{equation}
where $Pr(g|v)$ denotes the likelihood of $g$ in $v$, as given by the softmax outputs of a pre-trained object classification network. Such an approach has shown to be effective for unseen action classification~\citep{jain2015objects2action}. Here, we identify three additional semantic priors to improve both unseen action classification and localization.

\subsection{Priors for ambiguity, discrimination, and naming}
Similar to the common word embedding setup, for a video $v \in \mathcal{V}$, we seek to obtain a score for action $a \in \mathcal{A}$ using a set of global objects $\mathcal{G}$. Global objects generally come from deep networks~\citep{mettes2020shuffled} pre-trained on large-scale object datasets~\citep{deng2009imagenet}. We build upon current semantic matching approaches by providing three simple priors that deal with semantic ambiguity, non-discriminative objects, and object naming.
\\\\
\textbf{Object prior IV \oldreb{(semantic ambiguity prior)}:} \emph{A zero-shot likelihood estimation of action} a \emph{in video} v \emph{benefits from minimal semantic ambiguity between} a \emph{and global objects} $\mathcal{G}$.
\\\\
The score of a target action depends on the semantic relations to source objects. However, semantic relations can be ambiguous, since words can have multiple meanings depending on the context. For example for the action \emph{kicking}, an object such as \emph{tie} is deemed highly relevant, because one of its meanings is a draw in a football match~\citep{mettes2017spatial}. However, a \emph{tie} can also denote an entirely different object, namely a \emph{necktie}. Such semantic ambiguity may lead to the selection of irrelevant objects for an action.

To combat semantic ambiguity in the selection of objects, we consider two properties of object coherence across languages~\citep{malt1995}. First, most object categories are common across different languages. Second, the formation of some categories can nevertheless differ among languages. We leverage these two properties of object coherence across languages by introducing a multi-lingual semantic similarity. For computing multi-lingual semantic representations of words at a large-scale, we are empowered by recent advances in the word embedding literature, where embedding models have been trained and made publicly available for many languages~\citep{grave2018learning}.
%This supports the presence of a common underlying structure, which leads to a similar formation of object categories across cultures and languages. Second, the formation of some categories can nevertheless differ among cultures, societies, or individuals. A different categorization mainly arises due to a different point of view on the utility of objects, their symbolic value, or the level of knowledge of individuals. We leverage these two properties of object coherence across languages by introducing a multi-lingual semantic similarity. For computing multi-lingual semantic representations of words at a large-scale, we are empowered by recent advances in the word embedding literature, where embedding models have been trained and made publicly available for many languages~\citep{grave2018learning}.
In a multi-lingual setting, let $L$ denote the total number of languages to use. Furthermore, let $\tau_{l}(g)$ denote the translator for language $l \in L$ applied to object $g$. Multi-lingual unseen action classification can then be done by simply updating the semantic matching function to:
\begin{equation}
\Psi_L(g, a) = \frac{1}{L} \sum_{l=1}^{L} \cos(\phi_l(\tau_{l}(g)), \phi_l(\tau_{l}(a))),
\label{eq:language}
\end{equation}
where $\phi_l$ denotes the semantic word embedding of language $l$. The multi-lingual semantic similarity states that for a high semantic match between object and action, the pair should be of a high similarity across languages. In this manner, accidental high similarity due to semantic ambiguity can be addressed, as this phenomenon is factored out over languages.
%Recall the example of the action \emph{kicking} and the object \emph{tie}. Contrary to English, in French, a \emph{tie} game is an \emph{\'egalit\'e} and a \emph{necktie} is a \emph{cravate}. Such a clear distinction in another language dampens the similarity with the object \emph{tie}.
%Moreover, this formulation enforces the correct high similarity across languages. For example, consider now the action \emph{playing the cello} and the object \emph{cello}. There is a direct correspondence between \emph{cello} in English and its translation in French \emph{violoncelle}. Including another language confirms the similarity with the object \emph{cello}.
\\\\
\textbf{Object prior V \oldreb{(object discrmination prior)}:}  \emph{A zero-shot likelihood estimation of action} a \emph{in video} v \emph{benefits from knowledge about which objects in} $\mathcal{G}$ \emph{are suitable for action discrimination}.
\\\\
The second semantic prior is centered around finding discriminative objects. Only using semantic similarity to select objects ignores the fact that an object can be non-discriminative, despite being semantically similar. For example, for the action \emph{diving}, the objects \emph{person} and \emph{diving board} might both correctly be considered as semantically relevant. The object \emph{person} is however not a strong indicator for the action \emph{diving}, as this object is present in many actions. The object \emph{diving board} on the other hand is a distinguishing indicator, as it is not shared by many other actions.

To incorporate an object discrimination prior, we take inspiration from object taxonomies. When organizing such taxonomies, care must be taken to convey the most important and discriminant information~\citep{murphy2004big}. Here, we are searching for the most unique objects for actions, \ie objects with low inclusivity. It is desirable to select indicative objects, rather than focus on objects that are shared among many actions. To do so, we propose a formulation to predict the relevance of every object for unseen actions. We extend the action-object matching function as follows:
\begin{equation}
\Psi_r(g, a) = \Psi(g, a) + r(g, \cdot, a),
\label{eq:semsim-p2}
\end{equation}
where $r(g, \cdot, a)$ denotes a function that estimates the relevance of object $g$ for the action $a$. We propose two score functions. The first penalizes objects that are not unique for an action $a$:
\begin{equation}
r_a(g, A, a) = \Psi(g, a) - \max_{c \in A \setminus a} \Psi(g, c).
\end{equation}
An object $g$ scores high if it is relevant for action $a$ and for no other action. If either of these conditions are not met, the score decreases, which negatively affects the updated matching function.

The second score function solely uses inter-object relations for discrimination and is given as:
\begin{equation}
r_o(g, \mathcal{G}, a) = \Psi(g, a) - \frac{1}{|\mathcal{G}|} \sum_{g'\in \mathcal{G} \setminus g} \Psi(g, g')^{\frac{1}{2}}.
\end{equation}
Intuitively, this score function promotes objects that have an intrinsically high uniqueness across the set of objects, regardless of their match to actions. The square root normalization is applied to reduce the skewness of the object set distribution.
\\\\
\textbf{Object prior VI \oldreb{(object naming prior)}:} \emph{A zero-shot likelihood estimation of action} a \emph{in video} v \emph{benefits from a bias towards basic-level object names}.
\\\\
The third semantic prior concerns object naming. The matching function between actions and objects relies on the object categories in the set $\mathcal{G}$. The way objects are named and categorized has an influence on their matching score with an action. For example for the action \emph{walking with a dog}, it would be more relevant to simply name the object present in the video as a \emph{dog} rather than a \emph{domesticated animal}, or an \emph{Australian terrier}. Indeed, the \emph{dog} naming yields a higher matching score with the action \emph{walking with a dog} than the too generic \emph{domesticated animal} or too specific \emph{Australian terrier} namings.

As is well known, there exists a preferred entry-level of abstraction in linguistics, for naming objects~\citep{jolicoeur1984pictures,rosch1976}. The basic-level naming~\citep{rosch1976,rosch1988} is a trade-off between superordinates and subordinates. Superordinates concern broad category sets, while subordinates concern very fine-grained categories. Hence, basic-level categories are preferred because they convey the most relevant information and are discriminative from one another~\citep{rosch1976}. It would then be valuable to emphasize basic-level objects rather than objects from other levels of abstraction. Here, we enforce such an emphasis by using the relative WordNet depth of the objects in $\mathcal{G}$ to weight each object. Intuitively, the deeper an object is in the WordNet hierarchy, the more specific the object is and vice versa. To perform the weighting, we start from the beta distribution:
\begin{equation}
\begin{split}
\text{Beta}(d | \alpha, \beta) = & \frac{d^{\alpha-1} \cdot (1-d)^{\beta-1}}{B(\alpha,\beta)},\\
\quad B(\alpha,\beta) = & \frac{\Gamma(\alpha) \cdot \Gamma(\beta)}{\Gamma(\alpha+\beta)},\\
\end{split}
\end{equation}
where $d$ denotes the relative depth of an object and $\Gamma(\cdot)$ denotes the gamma function. Different values for  $\alpha$ and $\beta$ determine which levels to focus on. For a focus on basic-level we want to weight objects of intermediate level higher and the most specific and generic objects lower. We can do so by setting $\alpha = \beta = 2$. Setting $\alpha = \beta = 1$ results in the common setup where all objects are equally weighted. We incorporate the objects weights by adjusting the semantic similarity function between objects and actions.
\\\\
\textbf{Combined semantic priors.}
We combine the three semantic object priors into the following function of global objects for unseen actions:
\begin{equation}
\begin{split}
\ell_\text{video}(v, a) = \sum_{g \in \mathcal{G}_a} & ((\Psi_L(g, a) + \Delta(o,\cdot,a)) \cdot \\
& \text{Beta}(d_g | \alpha, \beta)) \cdot Pr(g|v),\\
\end{split}
\label{eq:vidscore}
\end{equation}
where $d_g$ denotes the depth of object $g$, [0,1] normalized based on the minimum WordNet depth (2) and maximum WordNet depth (18) over all objects in $\mathcal{G}$. In this formulation, the proposed embedding is more robust to semantic ambiguity, non-discriminative objects, and non-basic level objects compared to Equation~\ref{eq:semsim}.

\subsection{Object prior embedding}
%\cs{title of this section should be in line with comment in intro}
Unseen action localization and classification benefit from both a spatial and semantic priors. For unseen action localization, we obtain an object prior embedding by simply adding the tube score (Equation~\ref{eq:tubescore}) and the score of the corresponding video (Equation~\ref{eq:vidscore}). For unseen action classification we add the highest score of the tubes in the video with the video score.
%%%%%%%%%%%%%%%%%%%%%%%%%%%%%%%%%%%%%%%%%%%%%%%%%%
% SETUP
%%%%%%%%%%%%%%%%%%%%%%%%%%%%%%%%%%%%%%%%%%%%%%%%%%
\section{Experimental setup}
\label{sec:setup}

\subsection{Datasets}
We experiment on UCF Sports~\citep{rodriguez2008action}, J-HMDB~\citep{jhuang2013towards}, UCF-101~\citep{soomro2012ucf101}, Kinetics~\citep{carreira2017quo}, and AVA \citep{gu2018ava}. 
%
%\cs{Deze zin moet na toevoeging van Kinetics en AVA herschreven worden:}
%
Due to the lack of training examples, all these datasets still form open challenges in unseen action literature, even though high scores can be achieved with supervised approaches on \eg UCF-101~\citep{carreira2017quo,zhao2019dance}.
%\cs{statement why they are still very relevant for zero-shot?}

\textbf{UCF Sports} contains 150 videos from 10 actions such as \emph{running} and \emph{horse riding}~\citep{rodriguez2008action}. The videos are from sports broadcasts. We employ the test split provided by~\cite{lan2011discriminative}.

\textbf{J-HMDB} contains 928 videos from 21 actions such as \emph{brushing hair} and \emph{catching}~\citep{jhuang2013towards}, from HMDB~\citep{kuehne2011hmdb}. The videos focus on daily human activities. We employ the test split provided by~\cite{jhuang2013towards}.

\textbf{UCF-101} contains 13,320 videos from 101 actions such as \emph{skiing} and \emph{playing nasketball}~\citep{soomro2012ucf101}. The videos are taken from both sports and daily activities. We employ the test split provided by~\cite{soomro2012ucf101}.

%\cs{Voor kinetics en AVA ook kort iets zeggen over soort/bron van materiaal? kinetics vooral youtube, AVA vooral feature film?}

\textbf{Kinetics-400} contains 104,000 videos from 400 actions such as \emph{playing monopoly} and \emph{zumba}~\cite{carreira2017quo} from Youtube videos. We use all videos as test for unseen action classification.

\textbf{AVA}v2.2 contains 437 15-minutes clips from movies covering 80 atomic actions such as \emph{listening} and \emph{writing}~\cite{gu2018ava}. For 61 out of 64 validation videos, the YouTube links are still available and we use these as test videos for unseen action localization.

%\textbf{Hollywood2Tubes} contains 1,707 videos from 12 actions of Hollywood2~\citep{marszalek09}, which contain video clips from movies. The dataset is supplemented with spatio-temporal localization annotations by~\cite{mettes2019pointly}. We employ the test split provided by~\cite{marszalek09}.

Note that for all datasets, we exclude the use of \textit{any} information from the training videos. We employ the action labels and ground truth box annotations from the test videos to evaluate the zero-shot action classification and localization performance.

\subsection{Object priors sources}
\textbf{Object scores and detections.} To obtain person and local object box detections in individual frames, we employ Faster R-CNN~\citep{ren2015faster}, pre-trained on MS-COCO~\citep{lin2014microsoft}. The pre-trained network includes the person class and 79 objects, such as \emph{car}, \emph{chair}, and \emph{tv}. For the global object scores over whole videos, we apply a GoogLeNet~\citep{szegedy2015going}, pre-trained on 12,988 ImageNet categories~\citep{mettes2020shuffled}.
%, with a rate of two frames per second.
The object probability distributions are averaged over the sampled frames to obtain the global object scores.
\reb{On all datasets except AVA, frames are sampled at a fixed rate of 2 frames per second. On AVA, we use the annotated keyframes as frames. All frames have an input size of 224x224.}

\begin{table*}[t]
\centering
%\resizebox{0.8\linewidth}{!}{%
\begin{tabular}{ll@{\hskip 0.25in}ccccc@{\hskip 0.25in}ccccc}
\toprule
 & & \multicolumn{5}{c}{\textbf{Classification}} & \multicolumn{5}{c}{\textbf{Localization}}\\
 & & \multicolumn{5}{c}{Number of object detections} & \multicolumn{5}{c}{Number of object detections}\\
 &  & 0 & 1 & 2 & 5 & 10 & 0 & 1 & 2 & 5 & 10\\
\midrule
\textbf{Object prior I} & Person & 8.5 & - & - & - & - & 10.1 & - & - & - & -\\
\textbf{Object prior I+II} & + Objects & - & 21.3 & 19.2 & 27.7 & 27.7 & - & 22.8 & 22.8 & 24.4 & 23.6\\
\rowcolor{Gray}
\textbf{Object prior I+II+III} & + Spatial relations & - & 12.8 & 25.5 & \textbf{29.8} & \textbf{29.8} & - & 26.0 & 22.4 & \textbf{27.0} & 22.8\\
\bottomrule
\end{tabular}
%}
\caption{Effect of spatial object priors for unseen action classification (acc, \%) and localization (mAP@0.5, \%), on UCF Sports. We investigate three spatial prior settings; only person detections (I), person and object detections (I+II), and with the additional spatial prepositions between people and objects (I+II+III). For both unseen classification and localization, using the top five objects with spatial relations obtains the highest scores.}
\label{tab:study1}
\end{table*}

\textbf{Spatial priors sources.} For the spatial relations, we reuse the bounding box annotations of the training set of MS-COCO, as also used to pre-train the detection model, to obtain the prior prepositional knowledge between persons and objects.

\textbf{Semantic priors sources.}
For the semantic priors, we rely on FastText, pre-trained on 157 languages~\citep{grave2018learning}. This collection of word embeddings enables us to investigate multi-lingual semantic matching between actions and objects. For the multi-lingual experiments, we employ five languages: English, French, Dutch, Italian, and Afrikaans. We obtain action and object translations first from Open Multilingual WordNet~\citep{bond2013linking}. For the remaining objects and all actions, we use Google Translate with manual verification.

\textbf{Code.} The code is available at \url{https://github.com/psmmettes/object-priors-unseen-actions}.
%All code will be made publicly available.

%\cs{Are all the translations available?} \psmm{What do you mean here?} \cs{also, what about code?}

\subsection{Evaluation protocol}
We follow the zero-shot action evaluation protocol of \citep{jain2015objects2action,mettes2017spatial,zhu2018towards}, where no training is performed on a separate set of actions; the set of test actions are directly evaluated. For each dataset, we evaluate on the videos in the test set.
\reb{For classification experiments where the number of test actions is lower than the total number of actions in the dataset, we perform five random selections and report the mean accuracy and standard deviation.}
%For the classification experiments where we use fewer test actions than the total number of actions, we perform five random selections of test actions and report mean results with standard deviations.

For unseen action localization, we compute the spatio-temporal (st) overlap between action tube $a$ and ground truth $b$ from the same video as:
\begin{equation}
\text{st-iou}(a,b) = \frac{1}{|\Omega|} \sum_{f \in \Omega} \text{iou}_f(a,b), 
\end{equation}
where $\Omega$ states the union of frames in $a$ and $b$. The function $\text{iou}_f(a,b)$ is 0 if either one of the tubes is not present in frame $f$. For overlap threshold $\tau$, an action tube is positive if the tube is from a positive video, the overlap with a ground truth instances is at least $\tau$, and the ground truth instance has not been detected before.
\reb{For unseen action localization, we report the AUC and video mAP metrics on UCF Sports and J-HMDB, following~\cite{mettes2017spatial}. On AVA, we report frame mAP, following~\cite{gu2018ava}. Unless specified otherwise, the overlap threshold is 0.5.}
%We use mAP and AUC for evaluation, following the convention in literature.
For unseen action classification, we evaluate using multi-class classification accuracy.
%%%%%%%%%%%%%%%%%%%%%%%%%%%%%%%%%%%%%%%%%%%%%%%%%%
% RESULTS
%%%%%%%%%%%%%%%%%%%%%%%%%%%%%%%%%%%%%%%%%%%%%%%%%%
\section{Results}
\label{sec:results}

%%%%%%%%%%%%%%%%%%%%%%%%%%%%%%%%%%%%%%%%%%%%%%%%%%
% PART 1
%%%%%%%%%%%%%%%%%%%%%%%%%%%%%%%%%%%%%%%%%%%%%%%%%%
\subsection{Spatial object priors ablation}
In the first experiment, we evaluate the importance of spatial relations between persons and local object detections for unseen action classification and localization. We use the 80 local objects pre-trained on MS-COCO for this ablation study.
%\cs{Ik heb hier een zin over 80 coco objecten weggehaald, maar besef in exp 2 dat die niet weg kon.}
We investigate the desired number of local objects to select per action and the effect of modelling spatial relations.

\begin{figure}[t]
\centering
\begin{subfigure}{0.275\linewidth}
\includegraphics[width=\linewidth]{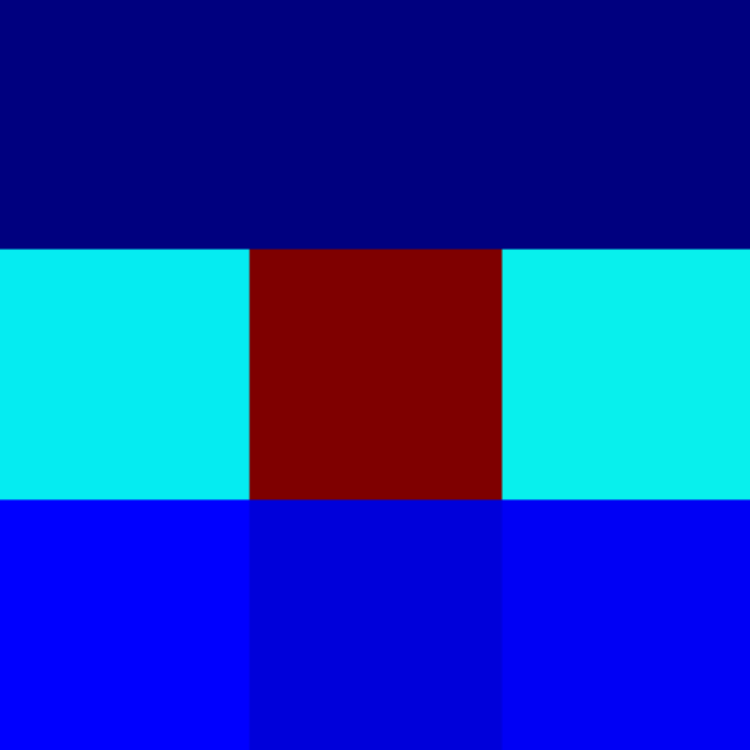}
\caption{Backpack.}
\end{subfigure}
\begin{subfigure}{0.275\linewidth}
\includegraphics[width=\linewidth]{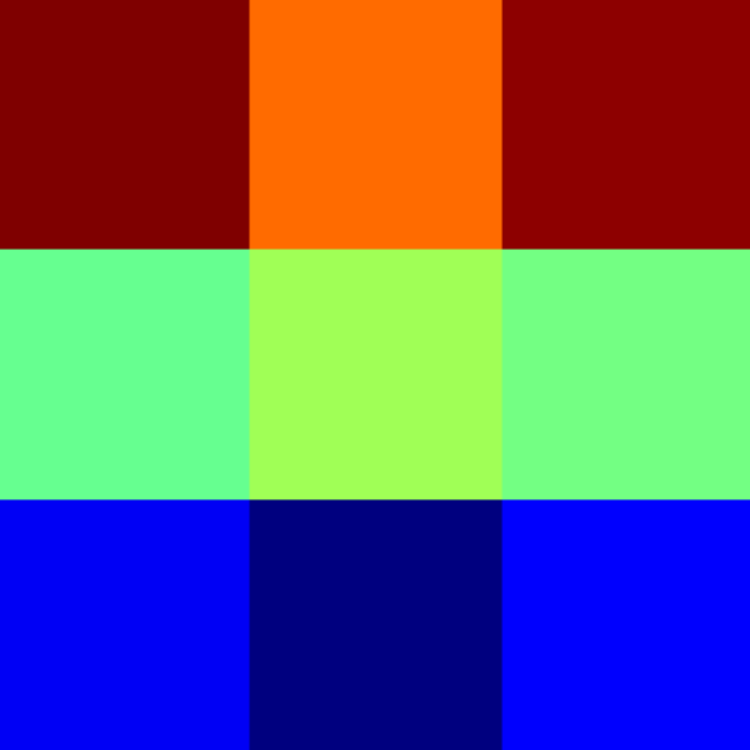}
\caption{Umbrella.}
\end{subfigure}
\begin{subfigure}{0.275\linewidth}
\includegraphics[width=\linewidth]{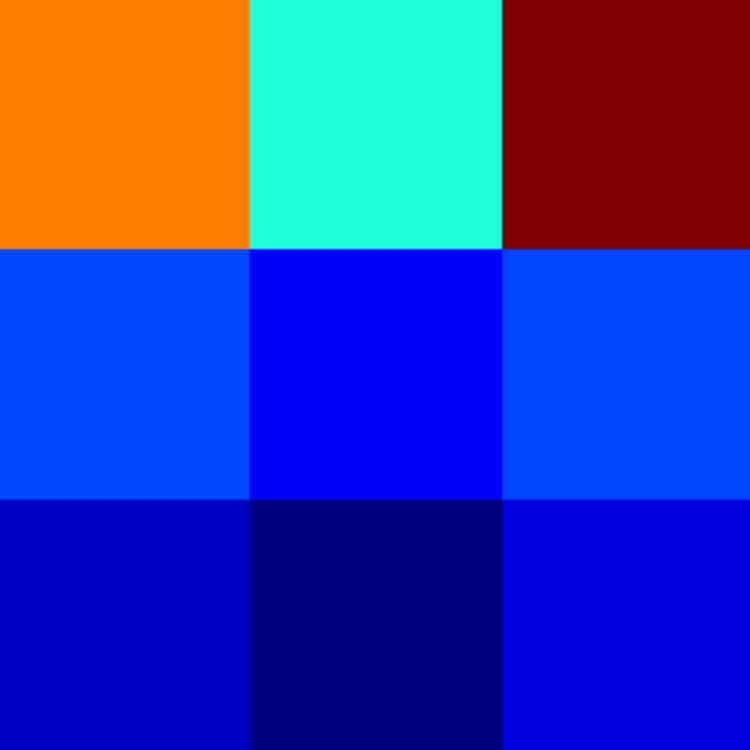}
\caption{Stop sign.}
\end{subfigure}
\\
\begin{subfigure}{0.275\linewidth}
\includegraphics[width=\linewidth]{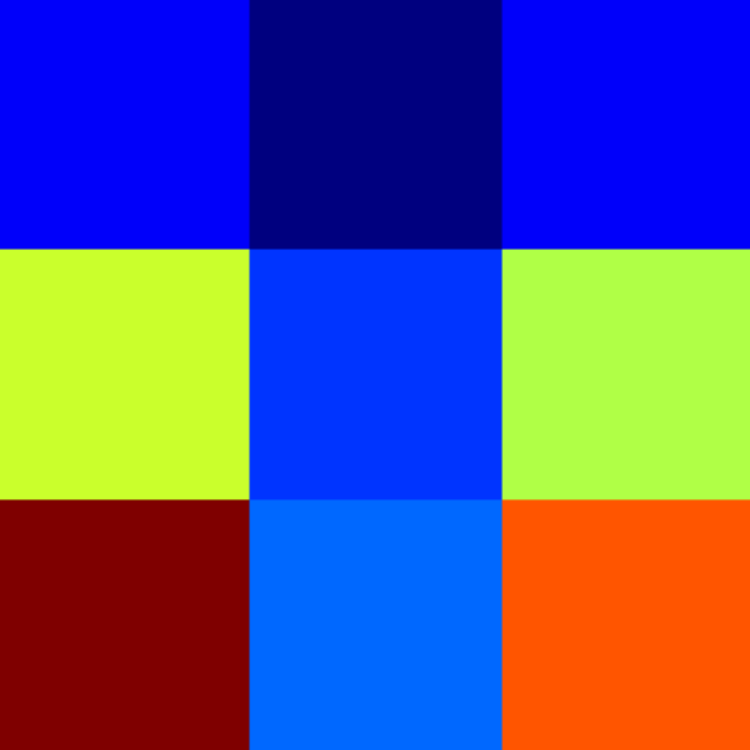}
\caption{Sheep.}
\end{subfigure}
\begin{subfigure}{0.275\linewidth}
\includegraphics[width=\linewidth]{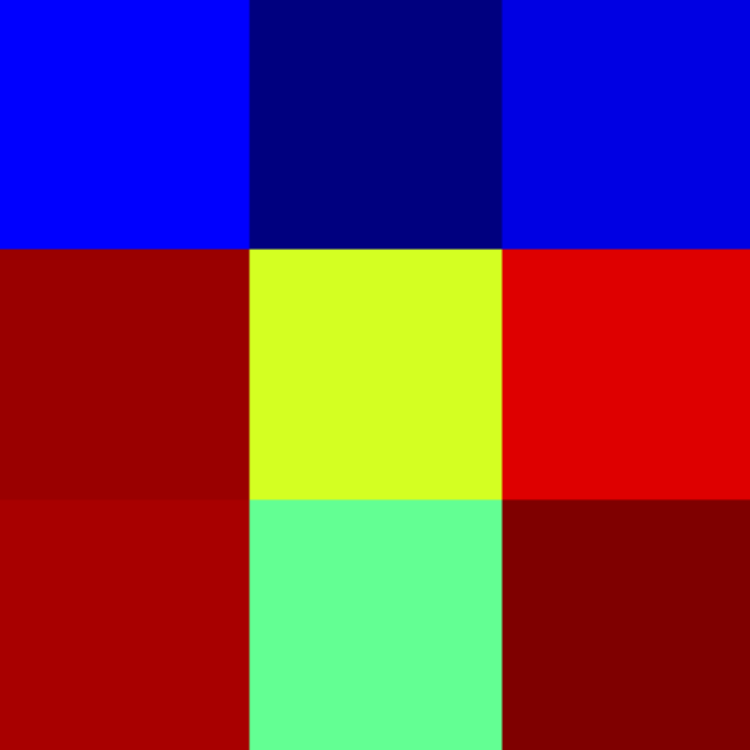}
\caption{Horse.}
\end{subfigure}
\begin{subfigure}{0.275\linewidth}
\includegraphics[width=\linewidth]{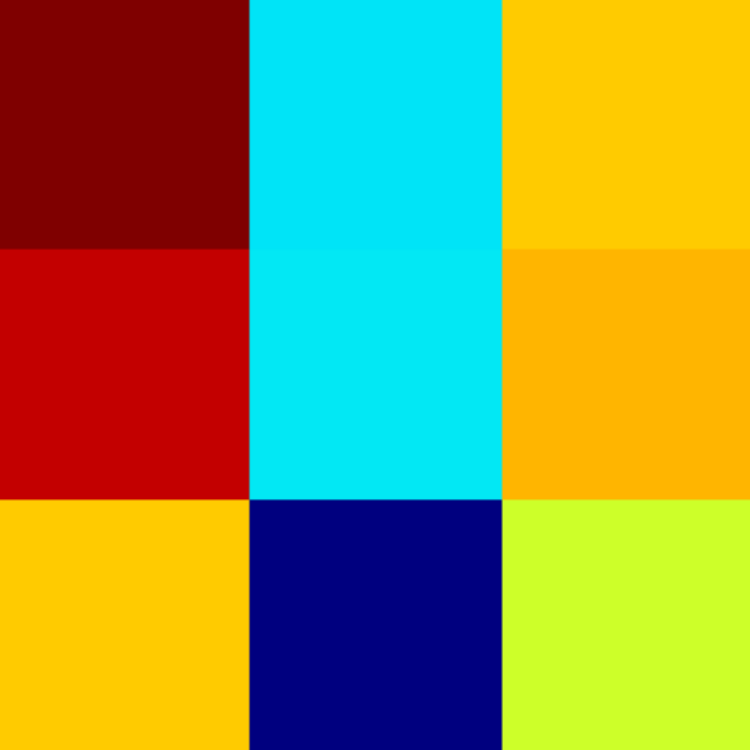}
\caption{Giraffe.}
\end{subfigure}
\caption{Spatial preposition priors for six local objects. Different objects have different spatial preferences relative to persons. These prepositional preferences align with our intuitions of the objects, \eg an umbrella tends to be above a person, while a backpack tends to be on a person.}
\label{fig:exemplars}
\end{figure}

Results are shown in Table~\ref{tab:study1}. When relying on only the first prior, person detections, we unsurprisingly obtain random classification and localization scores, since there is no direct manner to differentiate actions. Naturally, the first object prior is still vital, since it determines which boxes to consider in video frames. When adding the second prior, we find that the scores improve drastically for both classification and localization. Objects are indicative for unseen actions, whether actions need to be classified or localized.
%\cs{Ik denk dat je iets anders beoogt te zeggen dan je hier schrijft:}
%This result shows that objects are also important for actions in a localized setting.

%the three levels of awareness outlined for our proposed embedding: person detections only, person and object detections, and person and object detections along with spatial relations. First, we observe the importance of local objects themselves. When using person detections only, we unsurprisingly obtain random classification and localization performance, since there is no direct manner to differentiate actions.

%Upon incorporating local object detections, we find that the scores improve drastically for both classification and localization. This result shows that objects are also important for actions in a localized setting.
%
%\cs{Ik snap niet wat je met deze zin wilt zeggen:}
%This result reiterates the notion of relying on objects for unseen actions. 
%
Lastly, we include the spatial preposition prior. This provides a further boost in the results, showing that persons and objects have preferred spatial relations that can be exploited. In Figure~\ref{fig:exemplars}, we provide six discovered spatial relations from prior knowledge that are used in our action localization.

The results of Table~\ref{tab:study1} show that
%for classification, more objects result in better scores, both with and without the preposition prior. For localization there is no clear trend. High scores are obtained using one and five object per actions.
\oldreb{for unseen action classification, more local objects improve accuracy as they provide a richer source for action discrimination. For action localization, having many local objects may hurt, as the local box scoring becomes noisier, resulting in action tubes with lower overlap to the ground truth.}
Based on the scores obtained in this experiment, we recommend the use of spatial prepositions and five local object detections per action.

\begin{table}[t]
\centering
\begin{tabular}{lccc}
\toprule
 & \multicolumn{3}{c}{\textbf{UCF-101}}\\
 & \multicolumn{3}{c}{Number of test classes}\\
  & 25 & 50 & 101\\
\midrule
\rowcolor{Gray}
\textbf{Single language} & & & \\
English & 52.6 $\pm$ 4.7 & 43.3 $\pm$ 2.1 & 33.3\\
Dutch & 49.7 $\pm$ 3.5 & 40.4 $\pm$ 3.3 & 30.2\\
Portuguese & 44.4 $\pm$ 4.6 & 37.8 $\pm$ 4.0 & 28.5\\
Afrikaans & 43.6 $\pm$ 3.4 & 36.3 $\pm$ 4.0 & 27.7\\
French & 44.7 $\pm$ 2.5 & 36.2 $\pm$ 3.4 & 27.5\\
German & 38.0 $\pm$ 2.0 & 31.2 $\pm$ 2.6 & 26.0\\
\midrule
\rowcolor{Gray}
\textbf{Multi-lingual} & & & \\
English and Dutch & \textbf{54.3} $\pm$ \textbf{4.5} & \textbf{45.8} $\pm$ \textbf{2.4} & \textbf{35.7}\\
%Dutch and German & 46.1 $\pm$ 2.5 & 38.6 $\pm$ 2.3 & 30.3\\
%French and Portuguese & 44.8 $\pm$ 4.1 & 38.1 $\pm$ 2.3 & 29.3\\
All languages & 51.8 $\pm$ 4.8 & 43.2 $\pm$ 2.7 & 32.8\\
\bottomrule
\end{tabular}
\caption{Object prior IV \oldreb{(semantic ambiguity prior)} ablation. Unseen action classification accuracies (\%) on UCF-101 for multiple languages in the semantic matching for 25, 50, and 101 test classes. Combining two languages improves results. For this dataset, a combination of English and Dutch is best.}
\label{tab:exp1-s1}
\end{table}

\begin{figure}
\centering
\includegraphics[width=0.9\linewidth]{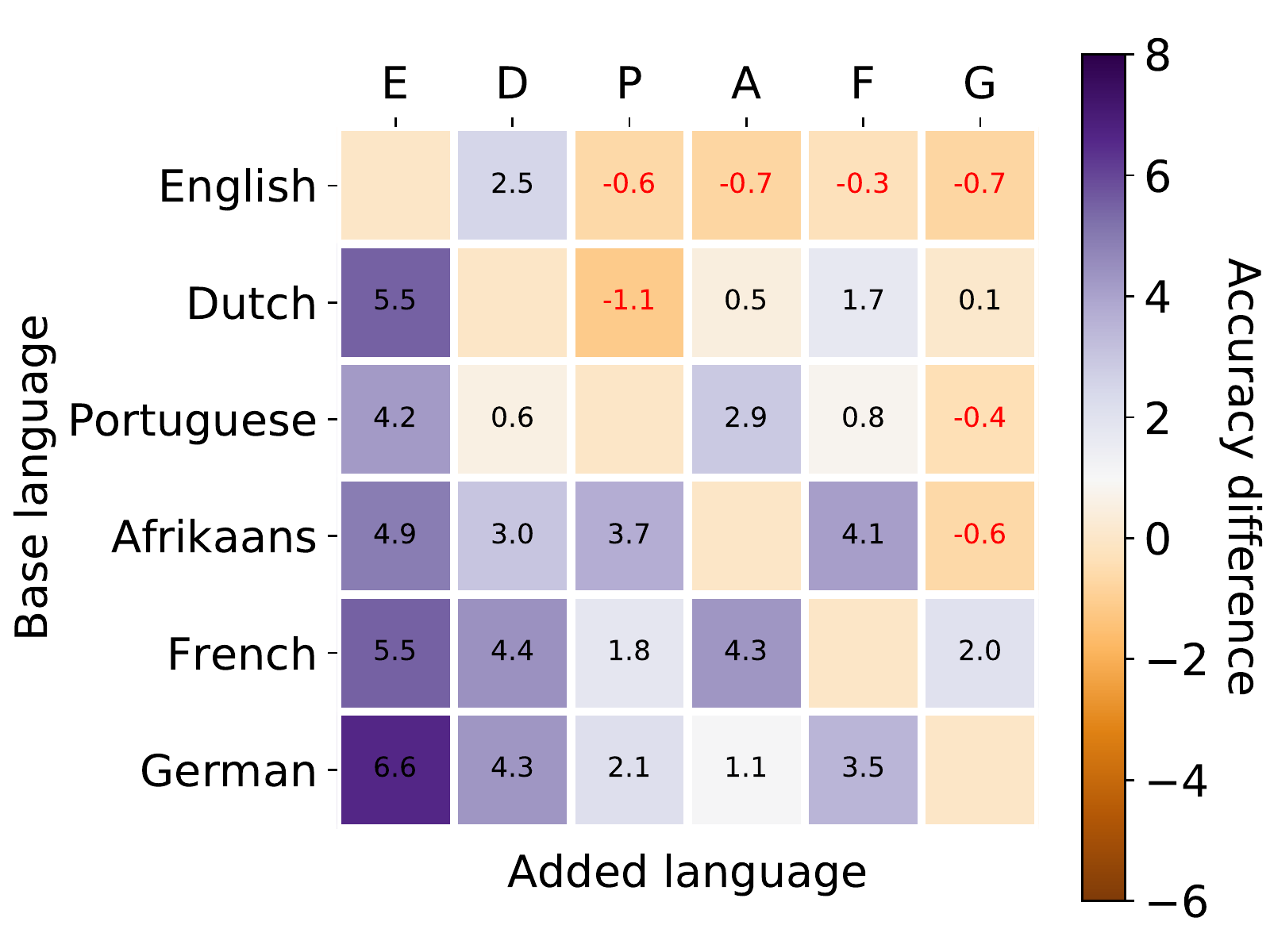}
\caption{\reb{Pairwise multilingual evaluation of all six languages on UCF-101 with all 101 test actions. The better the performance of the individual language, the more that language benefits others. For English, only adding the second best performing language (Dutch) is beneficial. When not taking English into account, we find that combining languages is mutually effective for seven out of the ten combinations.}}
\label{fig:prior-IV-pairwise}
\end{figure}

%%%%%%%%%%%%%%%%%%%%%%%%%%%%%%%%%%%%%%%%%%%%%%%%%%
% PART 2
%%%%%%%%%%%%%%%%%%%%%%%%%%%%%%%%%%%%%%%%%%%%%%%%%%
\subsection{Semantic object priors ablations}
In the second experiment, we perform ablation studies on the three semantic object priors for semantic matching between unseen actions and objects. We evaluate unseen action classification on UCF-101. Throughout this experiment, we focus on global object classification scores from the 12,988 ImageNet concepts applied and averaged over sampled video frames.
\\\\
\textbf{Object prior IV \oldreb{(semantic ambiguity prior)}.} We first investigate the importance of multi-lingual semantic similarity to deal with semantic ambiguity. We evaluate on three settings of UCF-101 for 25, 50, and 101 test classes. We perform this evaluation on all five individual languages, as well as their combination. We select the top-100 objects per action, following~\cite{mettes2017spatial}.

The results are shown in Table~\ref{tab:exp1-s1}. We first observe that individually English performs better than the other four languages. Dutch performs roughly three percent point lower, while the other three languages perform five to nine percent lower. A likely explanation for the lower results of the other languages is that the starting language of the objects and actions is English. The object and action names of the other languages are translated from English. Translation imperfections and breaking up compound nouns into multiple terms result in less effective word representations. As a result, there is a gap between English and the other languages.

\reb{
In Figure~\ref{fig:prior-IV-pairwise}, we show the relative accuracy scores for all language pairs on UCF-101 with all 101 test actions. We find that combining languages always boosts the least effective language of the pair. For the most effective English language, only the addition of Dutch results in a higher accuracy. For all other language pairs, the combined language performance is higher than the best individual language, except for German-Portuguese, German-Afrikaans, and Dutch-Portuguese. These are likely a result of poor individual performance (German) or low lexical similarity to other languages (Portuguese).} Overall, multi-lingual similarity with English and Dutch results in an improvement of 1.7\% 2.5\% and 2.4\% for 25, 50 and 101 classes. Further improvements are expected with better translations.
%\reb{
%When combining multiple languages, we find that the classification accuracy increases. There is no direct gain when using all languages. However, when restricting the multi-lingual action classification to the best two performing languages (English and Dutch), we find clear gains. Similarly, when two languages share the same origin, combining them can boost the scores, even when one language performs worse individually, e.g. high accuracies for 101 actions when adding German to Dutch and adding French to Portuguese. Multi-lingual similarity results in an improvement of 1.7\% 2.5\% and 2.4\% for 25, 50 and 101 classes.  Further improvements are expected with better translations.
%}
%When combining multiple languages, we find that the classification accuracy increases. The gain when using all languages is marginal for 50 or 101 test classes. However, when restricting the multi-lingual action classification to the best two performing languages (English and Dutch), we find larger gains. Multi-lingual similarity results in an improvement of 1.7\% 2.5\% and 2.4\% for respectively 25, 50 and 101 classes. Further improvements are expected with better translations.

\begin{figure}[t]
\centering
\includegraphics[width=0.9\linewidth]{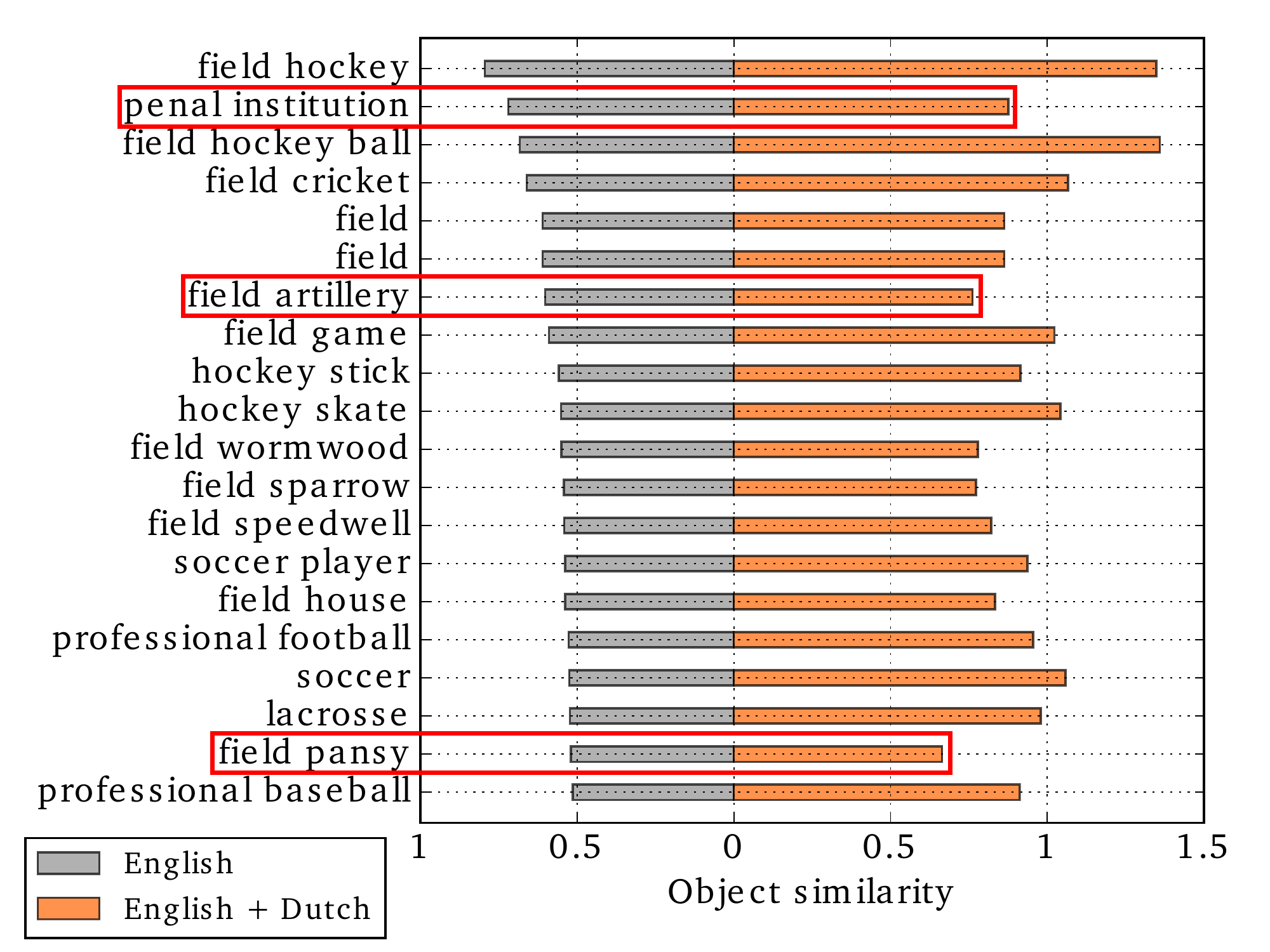}
\caption{Object prior IV \oldreb{(semantic ambiguity prior)} analysis with multiple languages for unseen recognition of the action \emph{field hockey penalty}. When relying on English, several irrelevant objects rank high due to semantic ambiguity (red boxes). When Dutch is added, ambiguous objects are downgraded, resulting in better recognition.}
\label{fig:exp1-s1-analysis}
\end{figure}

To investigate why multiple languages aid unseen action classification, we have performed a qualitative analysis for the action \emph{field hockey penalty} in UCF-101. We consider the most similar objects when using English only and when using English and Dutch combined. Figure~\ref{fig:exp1-s1-analysis} shows that for English only, several of the top ranked objects are not correct due to semantic ambiguity. These objects include penal institution, field artillery, and field wormwood. Evidently, such objects were selected because of their similarity to the English words field and penalty, but they are not related to the action of interest. When adding Dutch to the matching, such objects are ranked lower, because the ambiguity of these objects do not translate to Dutch. Hence, more relevant objects are ranked higher, which is also reflected in the results, where the accuracy increases from 0.07 to 0.27 for the action.

We conclude that using multiple languages for semantic matching between actions and objects reduces semantic ambiguity, resulting in improved unseen action classification accuracy.
\\\\
\textbf{Object prior V \oldreb{(object discrimination prior)}.}
For the object discrimination prior ablation, we investigate both the proposed object-based and action-based prior variants.
We again report on UCF-101 with 25, 50, and 101 test actions, with the top 100 objects selected per action.

\begin{table}[t]
\centering
\resizebox{\linewidth}{!}{%
\begin{tabular}{lccc}
\toprule
 & \multicolumn{3}{c}{\textbf{UCF-101}}\\
 & \multicolumn{3}{c}{Number of test classes}\\
  & 25 & 50 & 101\\
\midrule
Standard setup & 52.6 $\pm$ 4.7 & 43.3 $\pm$ 2.1 & 33.3\\
+ action-based discrimination & 53.2 $\pm$ 4.3 & 44.3 $\pm$ 1.9 & \textbf{34.3}\\
+ object-based discrimination & \textbf{54.0} $\pm$ \textbf{3.6} & \textbf{44.7} $\pm$ \textbf{2.1} & 34.0\\
\bottomrule
\end{tabular}
}
\caption{Object prior V \oldreb{(object discrimination prior)} ablation. Unseen action classification accuracy (\%) on UCF-101 with the proposed object discrimination functions for the English language. Both action-based and object-based discrimination aid recognition, especially when using fewer actions.}
\label{tab:exp1-s2}
\end{table}

%\begin{figure}[t]
%\centering
%\begin{subfigure}{0.457\linewidth}
%\includegraphics[width=\linewidth]{images/wordcloud-ApplyEyeMakeup.png}
%\caption{\emph{apply eye makeup.}}
%\end{subfigure}
%\begin{subfigure}{0.523\linewidth}
%\includegraphics[width=\linewidth]{images/wordcloud-PizzaTossing.png}
%\caption{\emph{pizza tossing.}}
%\end{subfigure}
%\caption{Object prior V \reb{(object discrimination prior)} analysis \reb{for two UCF-101 actions}. We visualize objects deemed most discriminative (gray) and least discriminative (red) for the actions a) \emph{apply eye makeup} and b) \emph{pizza tossing}. The proposed functions highlight objects that are both relevant and discriminative for the actions, purely by evaluating the semantic relations between objects and other objects or actions.}
%\label{fig:exp1-s2-analysis}
%\end{figure}

\begin{table}[t]
\centering
\resizebox{\linewidth}{!}{%
\begin{tabular}{rc|cl}
\toprule
\multicolumn{2}{c}{\textbf{Apply eye makeup}} & \multicolumn{2}{c}{\textbf{Pizza tossing}}\\
\midrule
\multicolumn{4}{c}{\textbf{Most discriminative objects}}\\
Makeup & \cellcolor{Gray}0.63 & \cellcolor{Gray}0.88 & Pizza\\
Eyeliner & \cellcolor{Gray}0.57 & \cellcolor{Gray}0.85 & Pepperoni pizza\\
Eyebrow pencil & \cellcolor{Gray}0.52 & \cellcolor{Gray}0.82 & Sausage pizza\\
Eyeshadow & \cellcolor{Gray}0.51 & \cellcolor{Gray}0.79 & Cheese pizza\\
Mascara & \cellcolor{Gray}0.50 & \cellcolor{Gray}0.78 & Anchovy pizza\\
\midrule
\multicolumn{4}{c}{\textbf{Least discriminative actions}}\\
Edam (cheese) & \cellcolor{Gray}-0.07 & \cellcolor{Gray}-0.06 & Argali (sheep)\\
Hokan (people) & \cellcolor{Gray}-0.07 & \cellcolor{Gray}-0.07 & Lhasa (terrier dog)\\
Lincoln (sheep) & \cellcolor{Gray}-0.07 & \cellcolor{Gray}-0.07 & Yautia (food)\\
Dicot (plant) & \cellcolor{Gray}-0.08 & \cellcolor{Gray}-0.08 & Caddo (people)\\
Loranthaceae (plant) & \cellcolor{Gray}-0.09 & \cellcolor{Gray}-0.08 & Filovirus (virus)\\
\bottomrule
\end{tabular}
}
\caption{
%\cs{waarom zijn getallen rood, en niet grijs net als in table 7? BTW wat zijn de getallen? AP? Misschien nog een header toevoegen boven elke column?} \psmm{Switched to gray, added most and least discriminative as headers, note about what the numbers are in the caption.}
\oldreb{Object prior V \oldreb{(object discrimination prior)} analysis for two UCF-101 actions. We show objects deemed most and least discriminative for \emph{apply eye makeup} and \emph{pizza tossing}, along with their scores. By finding out which objects are uniquely discriminative for an action in comparison to all other actions, we are able to highlight relevant objects and in turn improve unseen action classification.}}
%The proposed functions highlight objects that are both relevant and discriminative for the actions, purely by evaluating the semantic relations between objects and other objects or actions.}}
\label{tab:exp1-s2-analysis}
\end{table}

\begin{table}[t]
\centering
%\resizebox{\linewidth}{!}{%
\begin{tabular}{lccr}
\toprule
Weighting preference & $\alpha$ & $\beta$ & accuracy\\
\midrule
Uniform & 1 & 1 & 43.3 $\pm$ 2.1\\
Specific only & 5 & 1 & 7.6 $\pm$ 0.7\\
Generic only & 1 & 5 & 30.2 $\pm$ 1.1\\
\rowcolor{Gray}
Basic-level & 2 & 2 & \textbf{43.9} $\pm$ \textbf{2.0}\\
\bottomrule
\end{tabular}
%}
\caption{Object prior VI \oldreb{(object naming prior)} ablation. Unseen action classification accuracy (\%) on UCF-101 for 50 test classes using English. Only a small gain is feasible with a focus on basic-level objects compared to uniform weighting.}
\label{tab:exp1-s3}
\end{table}

\begin{table*}[t]
\centering
%\resizebox{\linewidth}{!}{%
\begin{tabular}{ccccccc}
\toprule
\multicolumn{3}{c}{\textbf{Object prior}} & \multicolumn{3}{c}{\textbf{Test actions}}\\
Semantic ambiguity (IV) & Object discrimination (V) & Object naming (VI) & 25 & 100 & 400\\
%IV & V & VI & 25 & 100 & 400\\
%{\scriptsize Semantic} & {\scriptsize Object} & {\scriptsize Object} &  &  & \\
%{\scriptsize Ambiguity} & {\scriptsize Discrimination} & {\scriptsize Naming} & & &\\
\midrule
& Random & & 4.0 & 1.0 & 0.3\\
& English only & & 21.8{\tiny$\pm$3.5} &  10.8{\tiny$\pm$1.0} & 6.0\\
\midrule
$\checkmark$ &  &  & 20.9{\tiny$\pm$4.1} & 10.8{\tiny$\pm$1.0} & 6.3\\
$\checkmark$ & $\checkmark$ & & 21.2{\tiny$\pm$3.9} & 10.7{\tiny$\pm$1.0} & 6.1\\
%\rowcolor{Gray}
$\checkmark$ & & $\checkmark$ & 22.0{\tiny$\pm$3.7} & 11.2{\tiny$\pm$1.0} & 6.4\\
$\checkmark$ & $\checkmark$ & $\checkmark$ & 21.9{\tiny$\pm$3.8} & 11.1{\tiny$\pm$0.8} & 6.4\\
\bottomrule
\end{tabular}
%}
\caption{
%\cs{Two-column van maken? Dan kun je naam van de priors voluit schrijven op 1 regel zonder te splitsen over drie regels. Midrule na English only regel?}
\reb{Unseen action classification on Kinetics-400 for the three semantic priors. Even with hundreds of unseen actions, the object priors help to assign action labels to videos. Across the three action sizes, semantic ambiguity and object naming work best, especially when having more unseen actions to choose from.}}
\label{tab:kinetics}
\end{table*}

The results in Table~\ref{tab:exp1-s2} show consistent improvements are obtained by both the action-based and the object-based variants. While the object-based taxonomy is preferred when recognizing 25 or 50 actions, the action-based taxonomy is preferred when recognizing 101 activities. In all three cases, incorporating a selection of the most discriminative objects yields better results. To highlight what kind of objects are boosted and subdued, we show the most and least discriminative objects of two actions in Table~\ref{tab:exp1-s2-analysis}.
\\\\
\textbf{Object prior VI \oldreb{(object naming prior)}.}
For the third semantic object prior, we evaluate the effect of weighting objects based on their WordNet depth to understand whether a bias towards basic-level objects is desirable in unseen action classification. This experiment is performed on UCF-101 for 50 test actions.

\begin{figure}[t]
\centering
\includegraphics[width=0.85\linewidth]{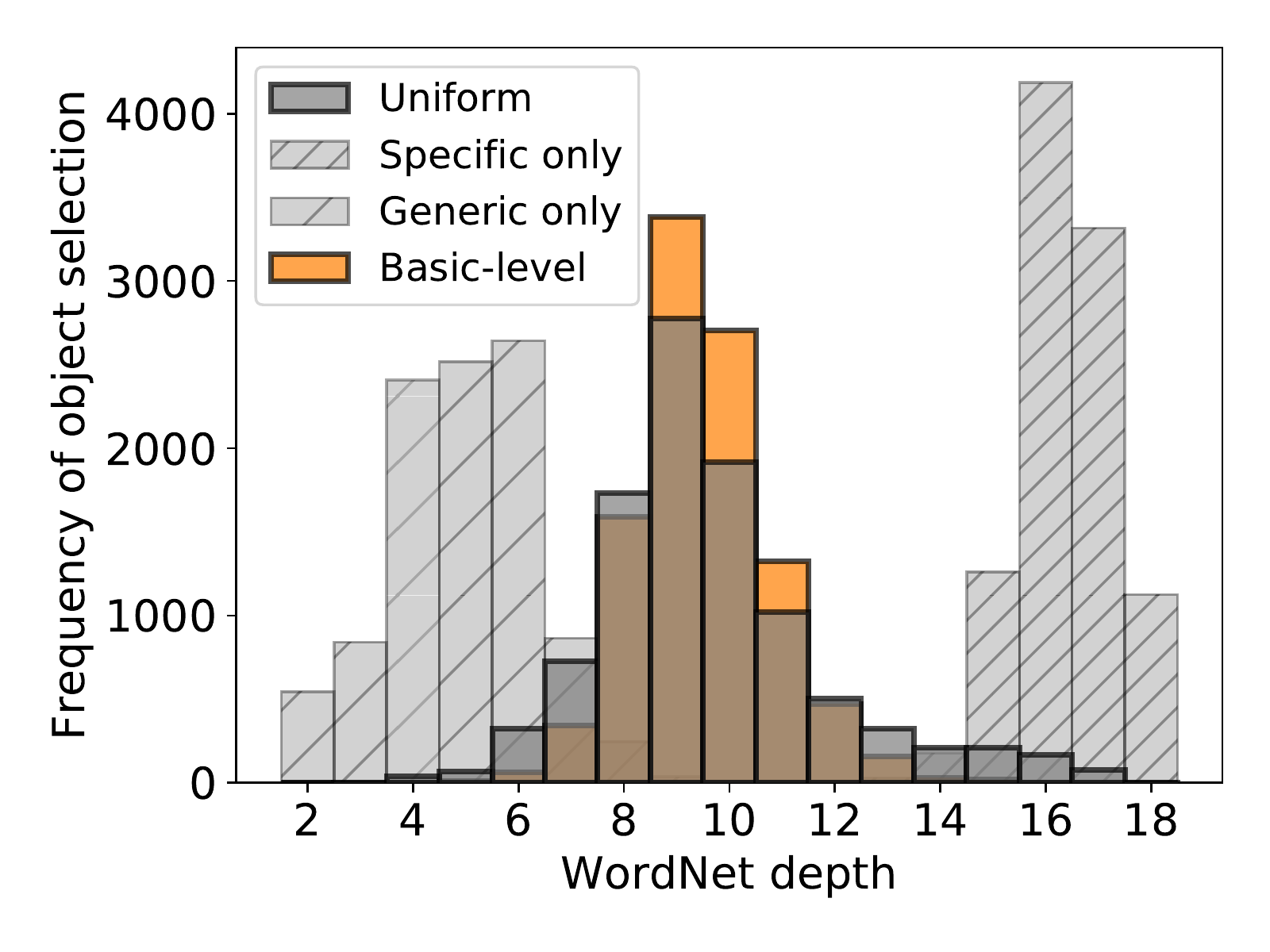}
\caption{Object prior VI analysis \oldreb{on UCF-101}. Akin to our basic-level object prior does the uniform weighting result in a distribution that favors basic-level objects. This explains the competitive performance of uniform weights versus the basic-level object prior; a bias towards basic-level objects is inherent in large-scale object sources. An explicit basic-level prior provides marginal gains.}
%The uniform weighting function (a) does not result in a uniform distribution, but in a distribution that favors basic-level objects. This explains the competitive performance of uniform weighting versus a basic-level object preference; a bias in basic-level objects is inherent in large-scale objects \cs{...?}, 
%
%\cs{deze zin loopt niet}
%reducing the need to focus on them explicitly.}
\label{fig:exp1-s3-analysis}
\end{figure}

%\begin{table}[t]
%\centering
%\resizebox{\linewidth}{!}{%
%\begin{tabular}{ccccccc}
%\toprule
%\multicolumn{3}{c}{\textbf{Object prior}} & \multicolumn{4}{c}{\textbf{Test %actions}}\\
%IV & V & VI & 25 & 50 & 100 & 400\\
%\midrule
%\multicolumn{3}{c}{Random} & 4.0 & 2.0 & 1.0 & 0.3\\
%$\checkmark$ &  &  & 20.9{\tiny$\pm$4.1} & 15.2{\tiny$\pm$1.9} & 10.8{\tiny$\pm$1.0} & 6.3\\
%$\checkmark$ & $\checkmark$ & & 21.2{\tiny$\pm$3.9} & 15.2{\tiny$\pm$1.5} & 10.7{\tiny$\pm$1.0} & 6.1\\
%\rowcolor{Gray}
%$\checkmark$ & $\checkmark$ & $\checkmark$ & \textbf{21.9}{\tiny$\pm$3.8} & \textbf{15.6}{\tiny$\pm$1.3} & \textbf{11.1}{\tiny$\pm$0.8} & \textbf{6.4}\\
%\bottomrule
%\end{tabular}
%}
%\caption{.}
%\label{tab:kinetics}
%\end{table}

\begin{figure*}[t]
\centering
\begin{subfigure}{0.25\textwidth}
\includegraphics[width=\textwidth]{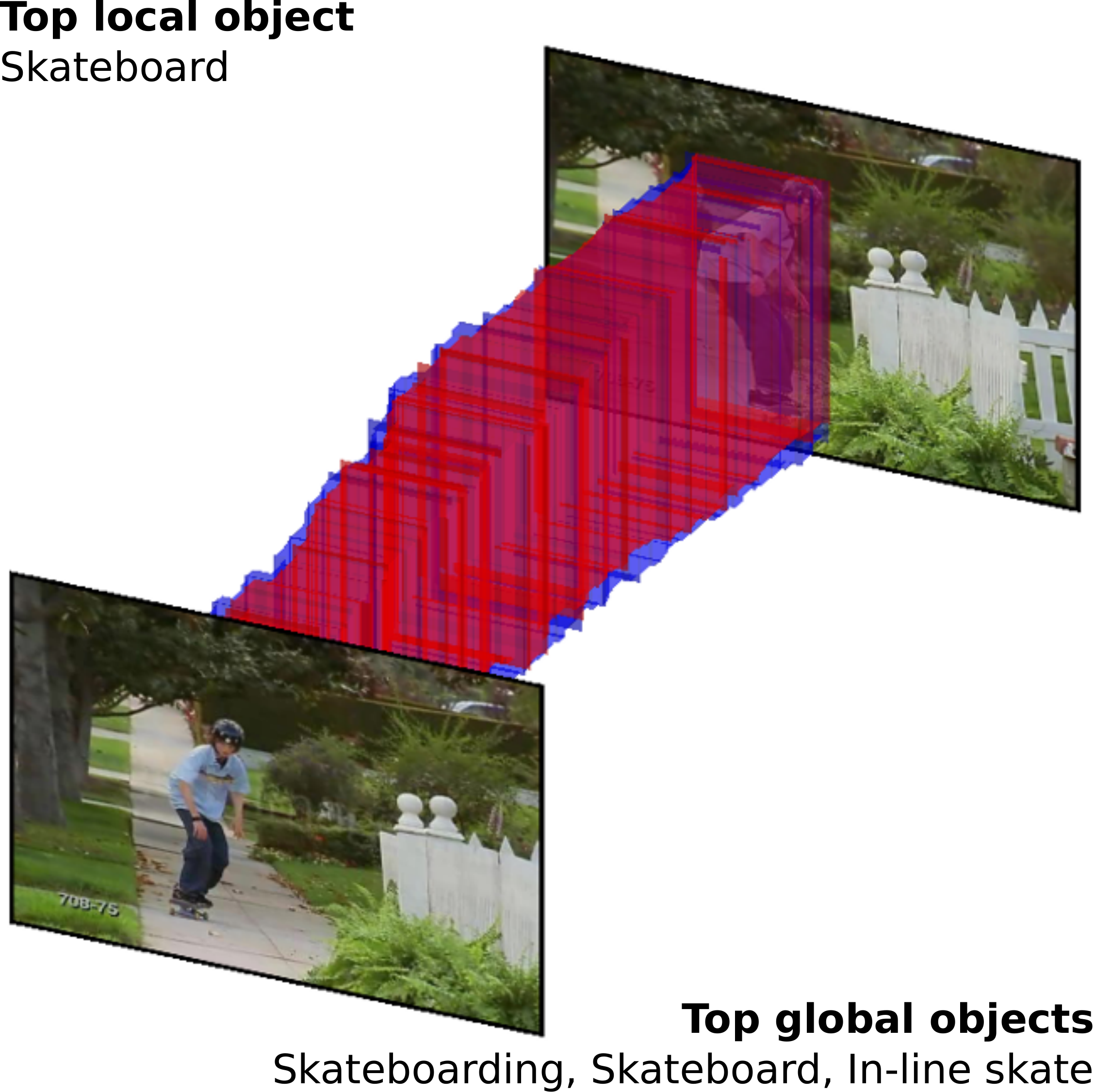}
\caption{Skateboarding.}
\end{subfigure}:
\hspace{0.75cm}
\begin{subfigure}{0.25\textwidth}
\includegraphics[width=\textwidth]{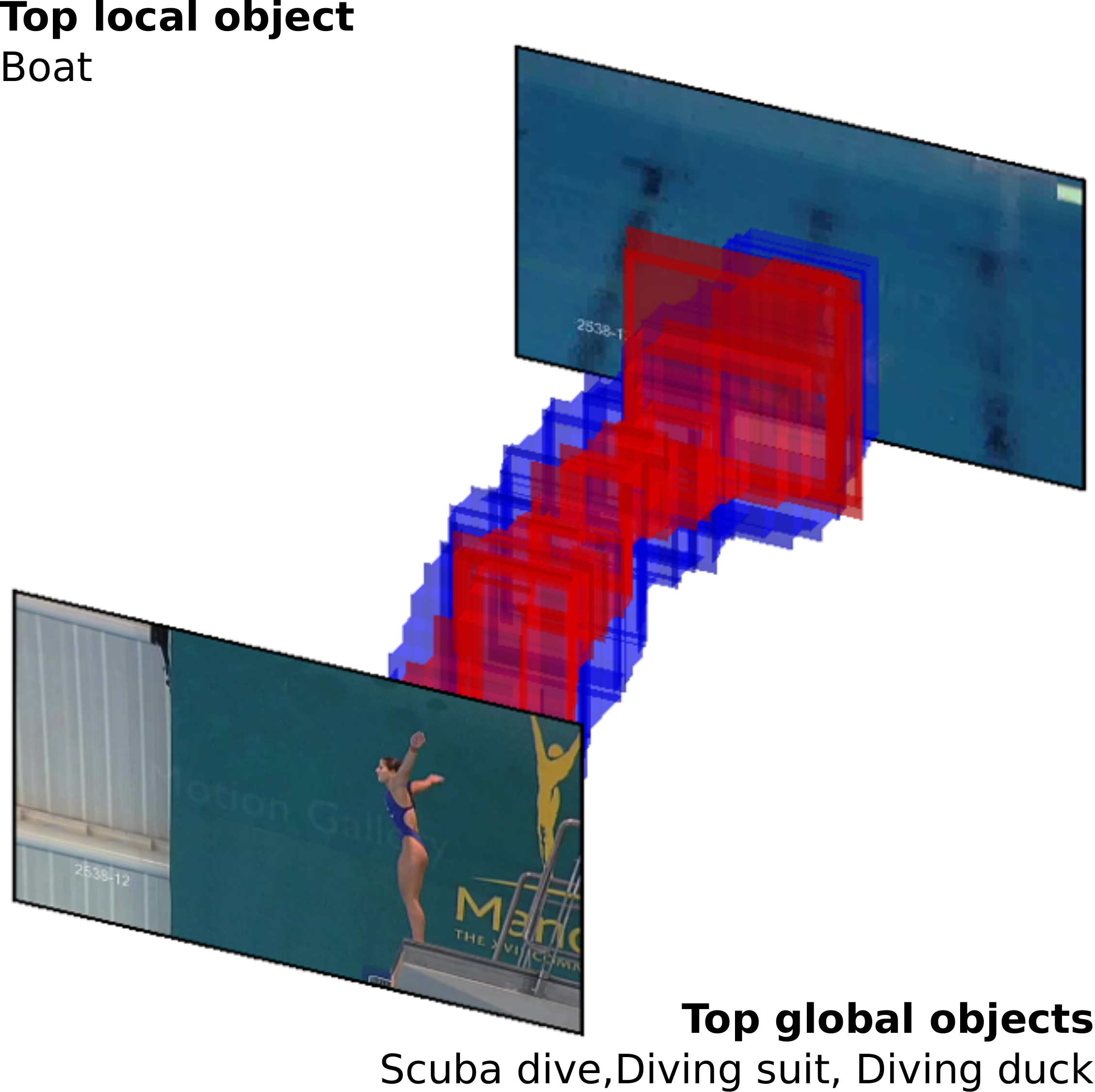}
\caption{Diving.}
\end{subfigure}
\hspace{0.75cm}
\begin{subfigure}{0.25\textwidth}
\includegraphics[width=\textwidth]{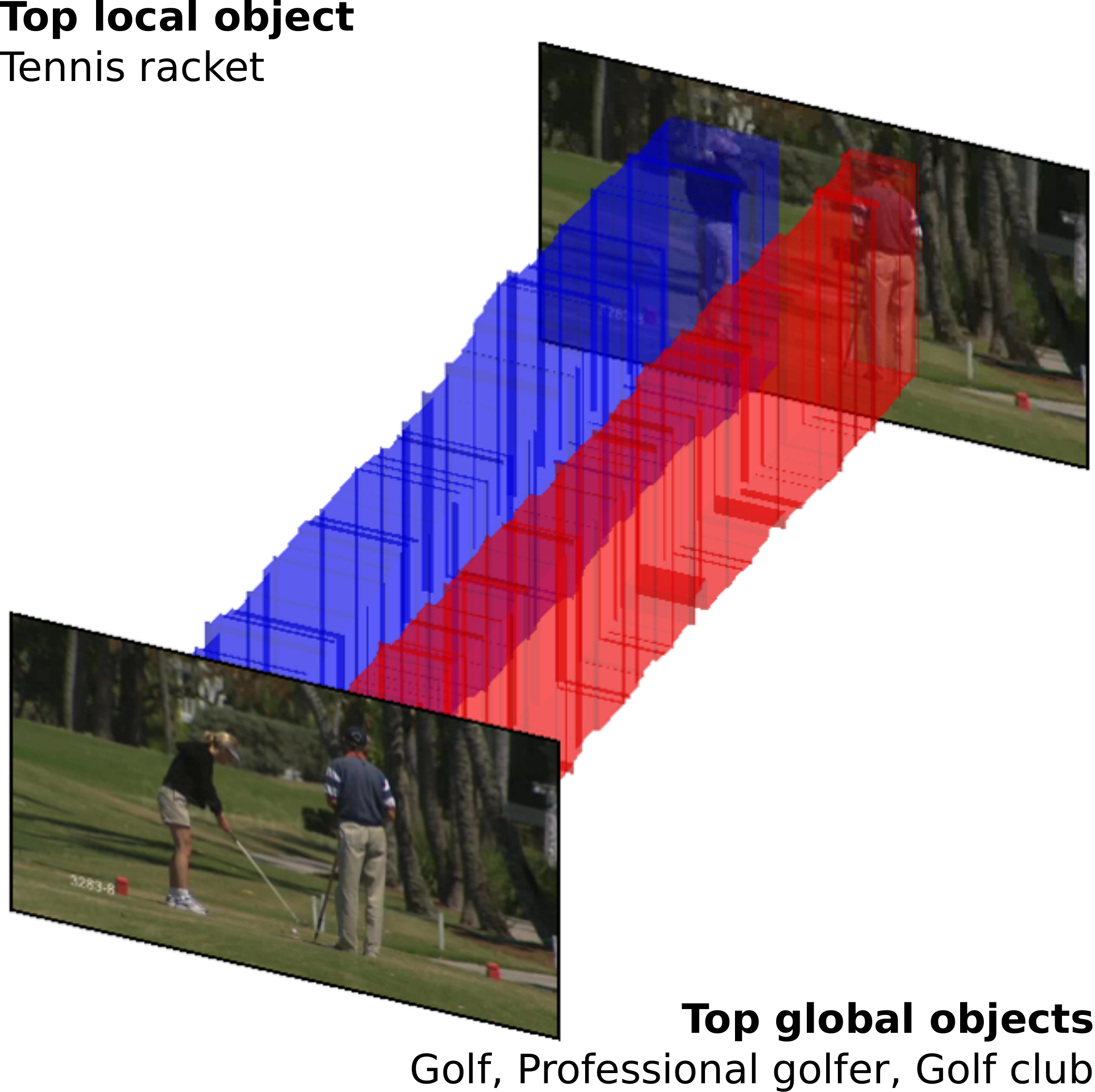}
\caption{Golfing.}
\end{subfigure}
\caption{Qualitative analysis on UCF Sports. For the video example of \emph{skateboarding} we obtain a correct localization due to a clear match with relevant objects.
%For the example of \emph{diving}, the example scores high for the action and the localization is roughly aligned with the ground truth, although the match with local and global objects is ambiguous. The most relevant high scoring objects are relevant for the action, but not present in the video. The action score is still high, since this is the only aquatic action in the dataset.
The example of \emph{golfing} obtains an incorrect localization. While the global objects are correct and relevant, the local object is incorrect. Upon inspection, we found that this error was due to the limited vocabulary of the local objects; no golf-based objects are present in MS-COCO.}
\label{fig:qual}
\end{figure*}

%\begin{figure}[t]
%\centering
%\includegraphics[width=0.825\linewidth]{images/ucf101-top-bottom-10.pdf}
%\caption{\reb{The top-10 and bottom-10 performing actions on UCF-101 using an English-Dutch vocabulary. Our approach is effective for actions with clear object interactions (\eg \emph{bowling, playing piano, horse riding,} and \emph{biking}). Actions can not be recognized when they are without direct object interactions (\eg \emph{lunges} and \emph{jumping jacks}) or when they use objects for which we have no detector or classifier (\eg \emph{yo yo} and \emph{ump rope}).}}
%\label{fig:ucf101-action-1}
%\end{figure}

Table~\ref{tab:exp1-s3} shows the results for the basic-level weighting preference compared to three baselines, \ie uniform (no preference), specific only, and generic only. We find that focusing on only the most specific or generic objects is not desirable and both result in a large drop in classification accuracy. The weighting preference for basic-level objects has a slight increase in accuracy compared to uniform, although the difference is small. This results shows that a prior for basic-level objects is not as effective as the semantic ambiguity and object discrimination priors.

To better understand our results, we have analysed the WordNet depth distribution of the top 100 selected objects for all actions in UCF-101. The distributions are visualized in Figure~\ref{fig:exp1-s3-analysis}. The two extreme preference weightings select objects from expected depth distributions and focus on the leftmost or rightmost side of the depth spectrum. Similarly for the basic-level weighting, objects from intermediate depth are selected. The uniform weighting however behaves unexpectedly and does not result in a uniform object depth distribution. In fact, this function also favors basic-level objects. The reason for this behaviour is found in the depth distribution of all 12,988 objects. For large-scale object collections, the WordNet depth distribution favors basic-level objects, following a normal distribution. As a result, the depth distribution of the selected objects follows a similar distribution, hence creating an inherent emphasis on basic-level objects. The basic-level object prior puts an additional emphasis on these kinds of objects and ignores specific and generic objects altogether.

\begin{figure}[t]
\centering
\includegraphics[width=0.775\linewidth]{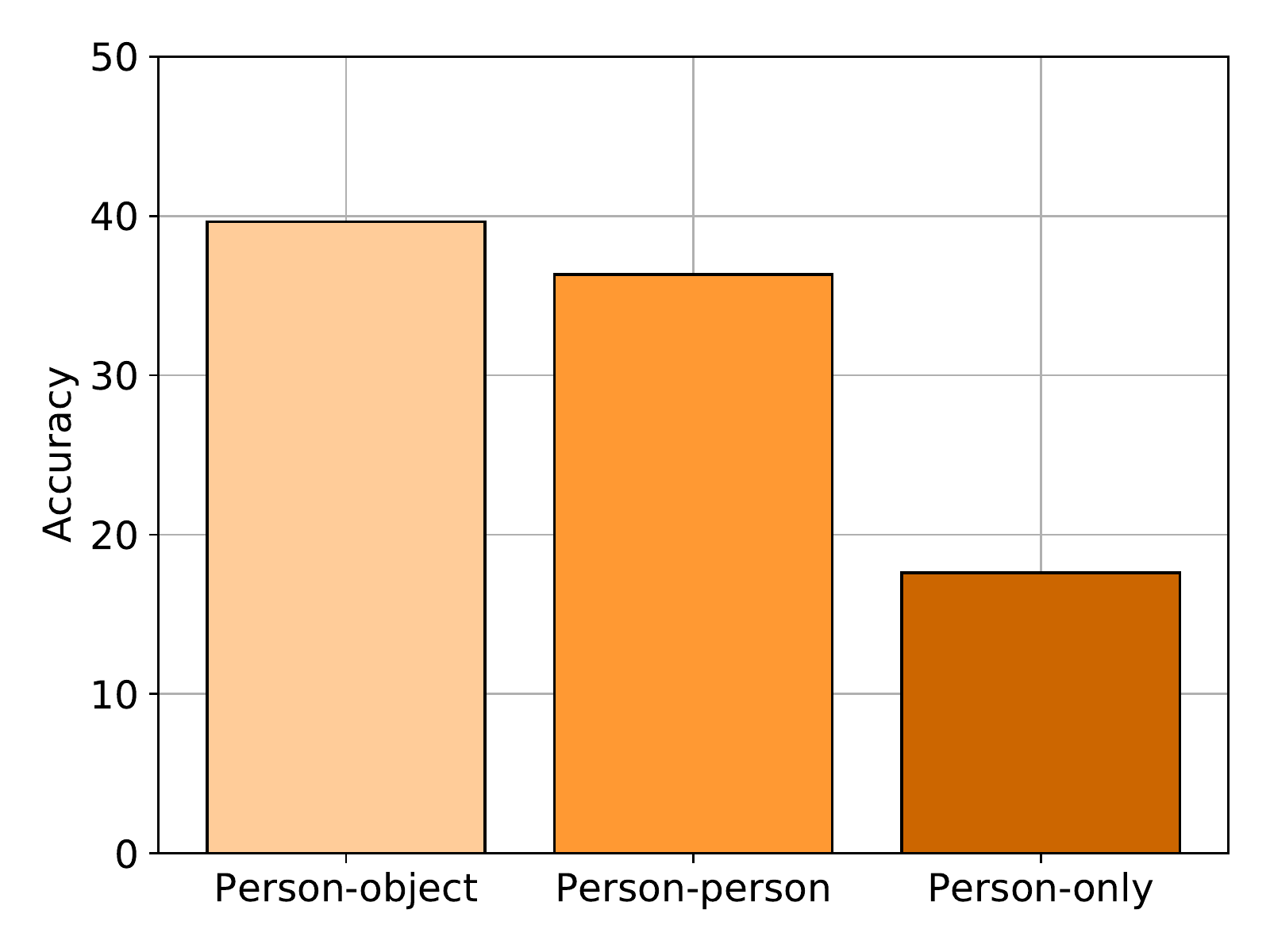}
\caption{\oldreb{UCF-101 accuracies aggregated into three categories; person-object, person-person, and person-only. As expected our approach favors person-object interactions due to our object priors. Person-only actions, such as gymnastics and fitness actions, obtain lower scores, highlighting the importance of having relevant objects to recognize actions in our approach.}}
\label{fig:ucf101-action-2}
\end{figure}

We conclude that a prior on basic-level objects \emph{is} important for unseen actions. Such a bias is inherently incorporated in large-scale object sources and no additional weighting is required to assist the object selection, although a small increase is feasible.
\\\\
\reb{\textbf{Combining semantic priors.} In Table~\ref{tab:kinetics}, we report the unseen action classification performance on Kinetics-400 using the semantic priors. Our approach does not require any class labels and videos during training, enabling a 400-way unseen action classification. When performing 400-way classification, the semantic ambiguity (IV) and object naming (VI) priors are most decisive, resulting in an accuracy of 6.4\%, compared to 0.25\% for random performance. For the Kinetics experiment, we evaluate unseen action classification as a function of the number of actions. For each size of the action vocabulary, we perform a random selection of the actions and perform 5 runs. We report both the mean and standard deviation.}
\\\\
\oldreb{\textbf{For which actions are semantic priors effective?} In Table~\ref{tab:top-bottom-actions}, we show respectively the top and bottom performing actions on UCF-101 and Kinetics when using our priors. \reb{On Kinetics, high accuracies can be achieved for actions such as \emph{playing poker} (65.0\%) and \emph{strumming guitar} (54.2\%), the accuracy is hampered by actions that can not be recognized, such as \emph{zumba} and \emph{situp}, likely due to the lack of relevant objects.} Figure~\ref{fig:ucf101-action-2} divides the UCF-101 actions into three classes; person-object, person-person, and person-only, to analyse when semantic priors are effective and when not.
}

\begin{table}[t]
\centering
\begin{tabular}{rc|cl}
\toprule
\multicolumn{2}{c}{\textbf{UCF-101}} & \multicolumn{2}{c}{\textbf{Kinetics}}\\
\midrule
%\multicolumn{4}{c}{\textbf{Top performing actions}}\\
Bowling & \cellcolor{Gray}98.1 & \cellcolor{Gray}65.0 & Playing poker\\
Ice dancing & \cellcolor{Gray}96.8 & \cellcolor{Gray}54.2 & Strumming guitar\\
Sumo Wrestling & \cellcolor{Gray}92.2 & \cellcolor{Gray}51.5 & Using segway\\
Horse riding & \cellcolor{Gray}91.5 & \cellcolor{Gray}48.5 & Golf chipping\\
Playing piano & \cellcolor{Gray}91.4 & \cellcolor{Gray}48.1 & Bowling\\
Playing sitar & \cellcolor{Gray}90.4 & \cellcolor{Gray}43.6 & Playing bass guitar\\
Rowing & \cellcolor{Gray}89.8 & \cellcolor{Gray}41.5 & Playing cymbals\\
Biking & \cellcolor{Gray}89.6 & \cellcolor{Gray}40.8 & Playing squash\\
Golf swing & \cellcolor{Gray}89.2 & \cellcolor{Gray}39.2 & Playing badminton\\
Playing violin & \cellcolor{Gray}88.0 & \cellcolor{Gray}39.2 & Playing cello\\
\midrule
%\multicolumn{4}{c}{\textbf{Bottom performing actions}}\\
Yo yo & \cellcolor{Gray}00.0 & \cellcolor{Gray}00.0 & Zumba\\
Hammer throw & \cellcolor{Gray}00.0 & \cellcolor{Gray}00.0 & Skiing\\
Jump rope & \cellcolor{Gray}00.0 & \cellcolor{Gray}00.0 & Egg hunting\\
Front crawl & \cellcolor{Gray}00.0 & \cellcolor{Gray}00.0 & Exercising arm\\
Frisbee catch & \cellcolor{Gray}00.0 & \cellcolor{Gray}00.0 & Exercise with ball\\
Floor gymnastics & \cellcolor{Gray}00.0 & \cellcolor{Gray}00.0 & Crosscountry skiing\\
Jumping jack & \cellcolor{Gray}00.0 & \cellcolor{Gray}00.0 & Faceplanting\\
Writing on board & \cellcolor{Gray}00.0 & \cellcolor{Gray}00.0 & Feeding fish\\
Lunges & \cellcolor{Gray}00.0 & \cellcolor{Gray}00.0 & Eating chips\\
Pizza tossing & \cellcolor{Gray}00.0 & \cellcolor{Gray}00.0 & Situp\\
\bottomrule
\end{tabular}
\caption{\reb{The top-10 and bottom-10 performing actions (acc, \%) on UCF-101 and Kinetics using an English-Dutch vocabulary. Across both datasets, our approach is effective for actions with clear object interactions (\eg \emph{bowling}, \emph{playing instruments}, \emph{horse riding}, and \emph{biking}). Actions can not be recognized when they are without direct object interactions (\eg fitness actions such as \emph{jumping jacks, zumba}, and \emph{exercising arm}) or when they use objects for which we have no detector or classifier (\eg \emph{yo yo} and \emph{exercising with exercise ball}).}}
\label{tab:top-bottom-actions}
\end{table}

\begin{table}[t]
\centering
\begin{tabular}{cccc}
\toprule
\multicolumn{2}{c}{\textbf{Object priors}} & \textbf{Classification} & \textbf{Localization}\\
spatial & semantic & accuracy (\%) & mAP@0.5 (\%)\\
\toprule
$\checkmark$ & & 29.8 & 27.0\\
 & $\checkmark$ & 59.6 & -\\
\rowcolor{Gray}
$\checkmark$ & $\checkmark$ & \textbf{68.1} & \textbf{34.9}\\
\bottomrule
\end{tabular}
\caption{Effect of combining spatial and semantic priors on the unseen action classification and localization results on UCF Sports. For both tasks, combining semantic matching for global objects with spatial matching for local objects is beneficial.}
\label{tab:exp2-s2}
\end{table}

\begin{figure*}[t]
\centering
\begin{subfigure}{0.25\textwidth}
\includegraphics[width=\textwidth]{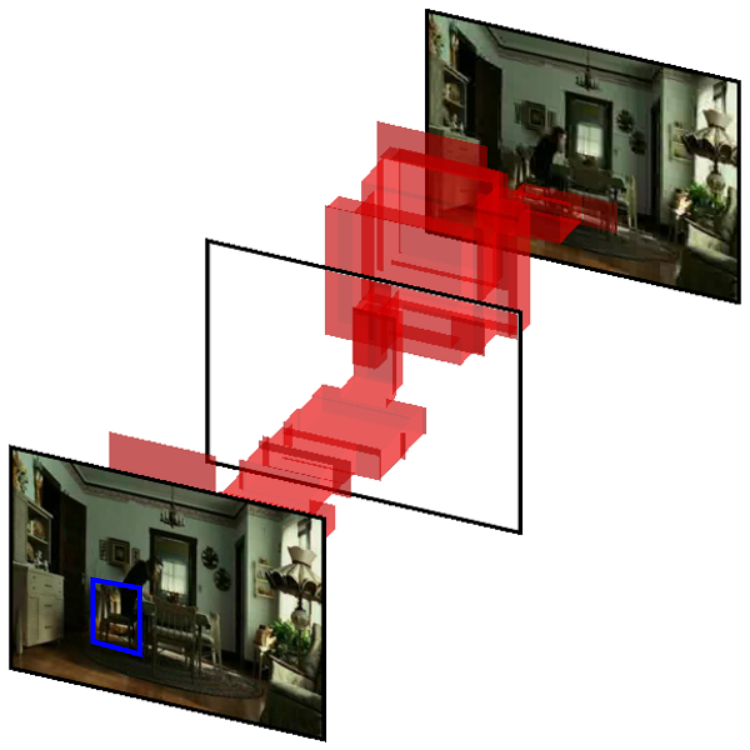}
\caption{Chair (bottom left).}
\end{subfigure}
\hspace{0.5cm}
\begin{subfigure}{0.25\textwidth}
\includegraphics[width=\textwidth]{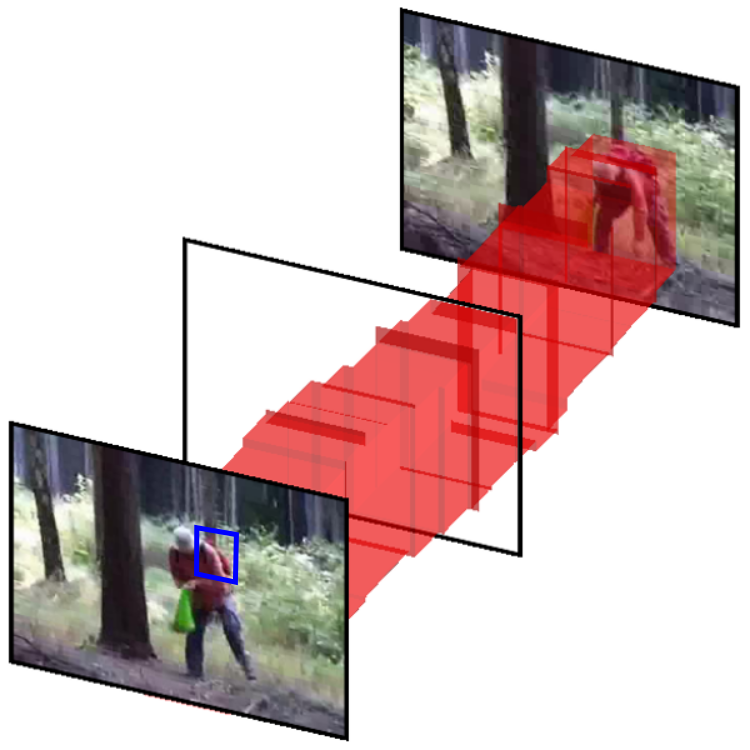}
\caption{Backpack (on).}
\end{subfigure}
\hspace{0.5cm}
\begin{subfigure}{0.25\textwidth}
\includegraphics[width=\textwidth]{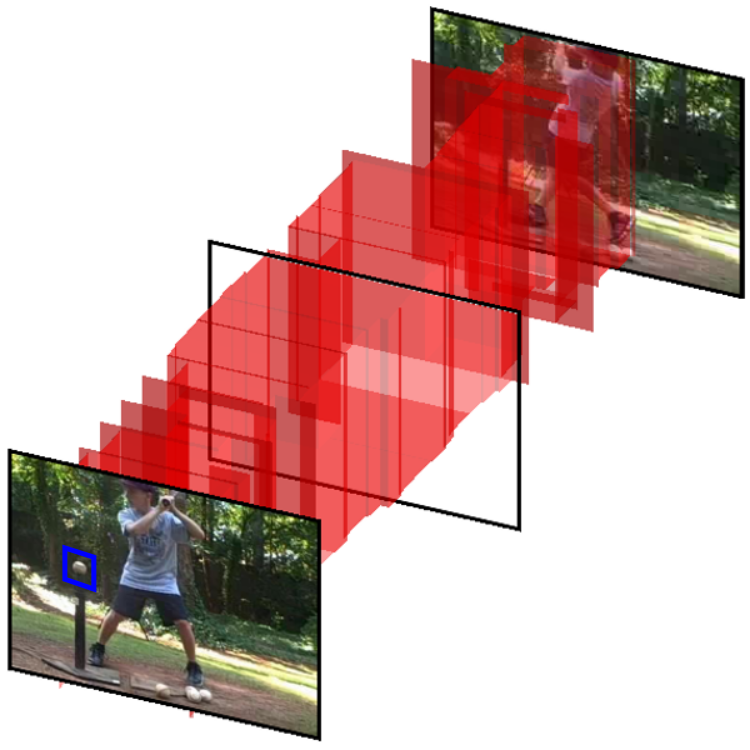}
\caption{Sports ball (left, size 0.1).}
\end{subfigure}
\caption{Qualitative results for action tube retrieval on J-HMDB. The examples for \emph{chair} and \emph{backpack} show that our object embedding is capable of retrieving relevant action locations from user queries on the fly. The example for \emph{sports ball} shows that we can additionally request a preferred object size. In this example, a localization with a baseball is retrieved, since a small ball size was queried.}
\label{fig:qual-retrieval}
\end{figure*}

%%%%%%%%%%%%%%%%%%%%%%%%%%%%%%%%%%%%%%%%%%%%%%%%%%
% PART 3
%%%%%%%%%%%%%%%%%%%%%%%%%%%%%%%%%%%%%%%%%%%%%%%%%%
\subsection{Combining spatial and semantic priors}
Based on the positive effect of the six individual spatial and semantic priors, we evaluate the impact on combining all priors for classification and localization. The results on UCF Sports are shown in Table~\ref{tab:exp2-s2}. Naturally, spatial objects priors are leading for unseen action localization, since this is impossible with semantic priors only. The reverse holds for action classification, where semantic priors on global objects are leading. We do find that for both tasks, using a combination of all priors is best. We recommend to use a combination of the six object priors to best deal with unseen actions.

\begin{figure*}[t]
\centering
\includegraphics[width=0.95\linewidth]{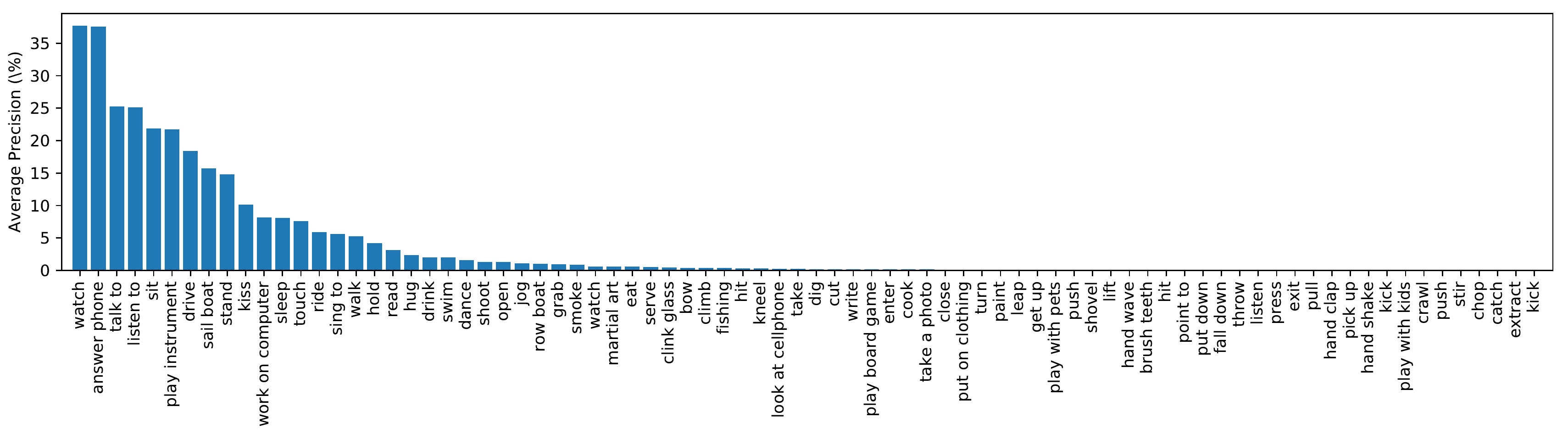}
\caption{
%\cs{A vertical bar plot with action class labels on the y-axis would be more clear and complete, can also be two columns. All reviewers ask for it.}
\reb{Quantitative results of our approach with all six object priors on AVA. We show the frame AP over all classes on the validation videos. The mean AP over all classes is 3.7\%, with notable high-performing actions that either involve clear interacting objects (\emph{answer phone}, \emph{player musical instrument}, and \emph{sail boat}) or involve multiple people that stand next to each other, in line with the spatial priors (\emph{listen to} and \emph{talk to a person}), highlighting that we can deal with multiple persons performing actions at the same time. Our approach struggles for single person actions, without any object interactions, such as \emph{crawl} and \emph{fall down}.}}
\label{fig:ava-perclass}
\end{figure*}

We show success and failure cases for unseen actions in Figure~\ref{fig:qual}. \oldreb{Adding the semantic priors on top of the spatial priors is especially beneficial when actions do not directly depend on an interacting object, see e.g. Figure~\ref{fig:qual}(b). Since there is no relevant interacting object for the \emph{diving} action, the corresponding tube relies solely on the person detection, resulting in a high overlap but with a low AP since the score is akin to non-\emph{diving} tubes. Adding the scores from the semantic priors however, results in the highest diving score for the shown action tube over all other test tubes. Interestingly, the global objects from the semantic priors are ambiguous for the action, \eg diving suit, but they still help for \emph{diving}, as it is the only aquatic action.}

%%Redundant: staat al in caption.
%A success case is shown on the left, with clearly correct local and global objects. An ambiguous case is shown in the middle, where a correct action was localized, although the most relevant objects do not precisely match with the video. A failure case is shown on the right, where the wrong person was localized due to a limited vocabulary for local objects.

%%%%%%%%%%%%%%%%%%%%%%%%%%%%%%%%%%%%%%%%%%%%%%%%%%
% PART 4
%%%%%%%%%%%%%%%%%%%%%%%%%%%%%%%%%%%%%%%%%%%%%%%%%%
\subsection{Action tube retrieval}
%\cs{Deze zin kan ik niet parsen:}
In the fourth experiment, we qualitative show the potential of our new task action tube retrieval.
%we perform a qualitative analysis of our new task of spatio-temporally retrieval unseen actions on-the-fly from user queries. 
%
In this setting, users query for desired objects, spatial prepositions, and optionally relative object size. In Figure~\ref{fig:qual-retrieval}, we show three example queries along with top retrieved action locations.

%%Redundant: staat al in caption.
%The example on the left and in the middle show that by specifying precise objects and spatial relations, we can obtain relevant action locations that are not pre-specified by the labels of the dataset. The example on the right shows that specifying a desired object size, a small object in this case, results in an action location in a relevant setting. We conclude from this experiment that our object embedding allows us to query objects, spatial relations, and object size to retrieve relevant action locations.

\begin{table}[t]
\centering
\resizebox{0.99\linewidth}{!}{%
\begin{tabular}{lrrr}
\toprule
 & \multicolumn{3}{c}{\textbf{UCF-101}}\\
 & \multicolumn{2}{c}{Number of actions} & accuracy\\
 &  train & test & \\
 \midrule
\cite{jain2015objects2action} & - & 101 & 30.3\\
\cite{mettes2017spatial} & - & 101 & 32.8\\
\rowcolor{Gray}
\emph{This paper} & - & 101 & 36.3\\
\cite{zhu2018towards} & 200 & 101 & 34.2\\
\cite{brattoli2020rethinking} & 664 & 101 & 37.6\\
\midrule
%\cite{kodirov2015unsupervised} & 51 & 50 & 14.0\\
\cite{mettes2017spatial} & - & 50 & 40.4\\
\rowcolor{Gray}
\emph{This paper} & - & 50 & 47.3\\
\cite{an2019spatiotemporal} & 51 & 50 & 17.3\\
\cite{bishay2019tarn} & 51 & 50 & 23.2\\
\cite{mishra2020zero} & 51 & 50 & 23.9\\
%\cite{li2016recognizing} & 51 & 50 & 26.8\\
\cite{mandal2019out} & 51 & 50 & 38.3\\
\cite{zhu2018towards} & 200 & 50 & 42.5\\
\cite{brattoli2020rethinking} & 664 & 50 & 48.0\\
\midrule
\cite{mettes2017spatial} & - & 20 & 51.2\\
\rowcolor{Gray}
\emph{This paper} & - & 20 & 61.1\\
%\cite{kodirov2015unsupervised} & 81 & 20 & 22.5\\
\cite{gan2016learning} & 81 & 20 & 31.1\\
\cite{bishay2019tarn} & 81 & 20 & 42.7\\
\cite{zhu2018towards} & 200 & 20 & 53.8\\
\bottomrule
\end{tabular}
}
\caption{
%\cs{Kunnen we Zhu en Brattoli niet beter beetje apart zetten? Bijv. elke keer onder This paper. Of sortering aanpassen obv train actions. This paper zou onder (of boven) Mettes en Snoek moeten staan, want dat is de meest eerlijke vergelijking. Ik zou de footnote tekens weghalen en in de caption zetten dat Zhu (200) ActivityNet actions gebruikt en Brattoli (664) Kinetics actions}
Comparison for unseen action classification accuracy (\%) on UCF-101 for multiple numbers of test classes. The train and test columns denote the number of action used for training and testing. Our approach is state-of-the-art in the unseen setting, where no training actions are used, and competitive to \cite{zhu2018towards} and \cite{brattoli2020rethinking}, who require extensive training on ActivityNet and Kinetics respectively.
%\oldreb{$\dagger$ and $\star$ denote paper with additional training on respectively ActivityNet and Kinetics.
%Across all settings, our approach is competitive to the state-of-the-art without the need for any videos to train on.
}
%For all settings, our approach obtains state-of-the-art results, without the need for examples during training.}
\label{tab:sota-classification}
\end{table}

\begin{table*}[t]
\centering
\resizebox{0.9\linewidth}{!}{%
\begin{tabular}{lrrrrrrrrrr}
\toprule
 & \multicolumn{5}{c}{\textbf{AUC}} & \multicolumn{5}{c}{\textbf{mAP}}\\
 \cmidrule(lr){2-6} \cmidrule(lr){7-11}
 & 0.1 & 0.2 & 0.3 & 0.4 & 0.5 & 0.1 & 0.2 & 0.3 & 0.4 & 0.5\\
%\midrule
 \cmidrule(lr){1-1} \cmidrule(lr){2-6} \cmidrule(lr){7-11}
\rowcolor{Gray}
\textbf{UCF Sports} & & & & & & & & & &\\
\cite{jain2015objects2action} & 38.8 & 23.2 & 16.2 & 9.9 & 7.2 & - & - & - & - & -\\
\cite{mettes2017spatial} & 43.5 & 39.3 & 37.1 & 35.7 & 31.1 & 47.4 & 43.5 & 42.1 & 32.0 & 23.2\\
\textit{This paper} & \textbf{47.3} & \textbf{43.0} & \textbf{40.7} & \textbf{37.9} & \textbf{33.1} & \textbf{61.2} & \textbf{54.2} & \textbf{54.0} & \textbf{41.5} & \textbf{34.9}\\
%\midrule
\cmidrule(lr){1-1} \cmidrule(lr){2-6} \cmidrule(lr){7-11}
\rowcolor{Gray}
\textbf{J-HMDB} & & & & & & & & & &\\
\cite{mettes2017spatial} & 34.6 & 33.3 & 30.5 & 26.8 & 23.0 & 27.5 & 27.0 & 23.2 & 19.2 & 15.1\\
\textit{This paper} & \textbf{37.3} & \textbf{37.1} & \textbf{33.9} & \textbf{31.0} & \textbf{26.7} & \textbf{32.1} & \textbf{31.5} & \textbf{27.2} & \textbf{22.6} & \textbf{17.6}\\
\bottomrule
\end{tabular}
}
\caption{
\oldreb{Unseen action localization comparisons on UCF Sports and J-HMDB using AUC and mAP across 5 overlap thresholds. Across all settings, we obtain improved results, indicating the effectiveness of our approach.}}
%Unseen action localization comparisons to~\cite{mettes2017spatial} on UCF Sports using AUC and mAP as metrics and J-HMDB using AUC. Across all comparisons, we obtain improved results, indicating the effectiveness of our approach.}
\label{tab:sota-localization-2}
\end{table*}

%%%%%%%%%%%%%%%%%%%%%%%%%%%%%%%%%%%%%%%%%%%%%%%%%%
% PART 5
%%%%%%%%%%%%%%%%%%%%%%%%%%%%%%%%%%%%%%%%%%%%%%%%%%
\subsection{Comparative evaluation}
In the fifth experiment, we compare our proposed approach to other works in action classification and localization without examples.  For the classification comparison, we report on the UCF-101 dataset, since it is most used for this setting. For the localization comparison, we report on the other two datasets. For all comparisons, we use both spatial and semantic object priors.
\\\\
%\cs{Je zegt niets over Zhu en Brattoli}
\textbf{Unseen action classification.} In Table~\ref{tab:sota-classification}, we show the unseen classification accuracies on UCF101 for three common dataset splits using 101, 50, and 20 test classes. We first note the difference in scores with our conference version~\citep{mettes2017spatial}, which are due to the three new semantic object priors. \reb{In the unseen setting, where no training actions are used, we are state-of-the-art. Moreover, we are competitive with zero-shot approaches that require extensive training on large-scale action datasets, such as \cite{zhu2018towards} and \cite{brattoli2020rethinking}.}
%We further find that across all settings, our approach outperforms alternatives. 
%
%\cs{Deze p.p. conventie is bedacht door Pascal Mettes, zou ik niet doen.}
%
%The difference to the state-of-the-art approach of~\cite{zhu2018towards} is 2.1 \% (absolute) for 101 test actions, 4.8 \% for 50 test actions, and 7.3 \% for 20 test actions. Note that compared to a number of other works, we do not require any seen actions in a training stage.
%We also note that
Each approach employs different prior knowledge, making a direct comparison difficult. The comparison serves to highlight the overall effectiveness of our approach.
%\begin{figure}[t]
%\centering
%\includegraphics[width=0.9\linewidth]{images/kinetics-zeroshot-numbers.pdf}
%\caption{\reb{Quantitative evaluation of unseen action recognition on Kinetics-400. The results show that performing unseen action recognition without access to any training videos is possible from object priors, although there is ample room for further improvement.}}
%\label{fig:kinetics}
%\end{figure}
\begin{figure*}[t]
\centering
\begin{subfigure}{0.265\textwidth}
\includegraphics[width=\textwidth]{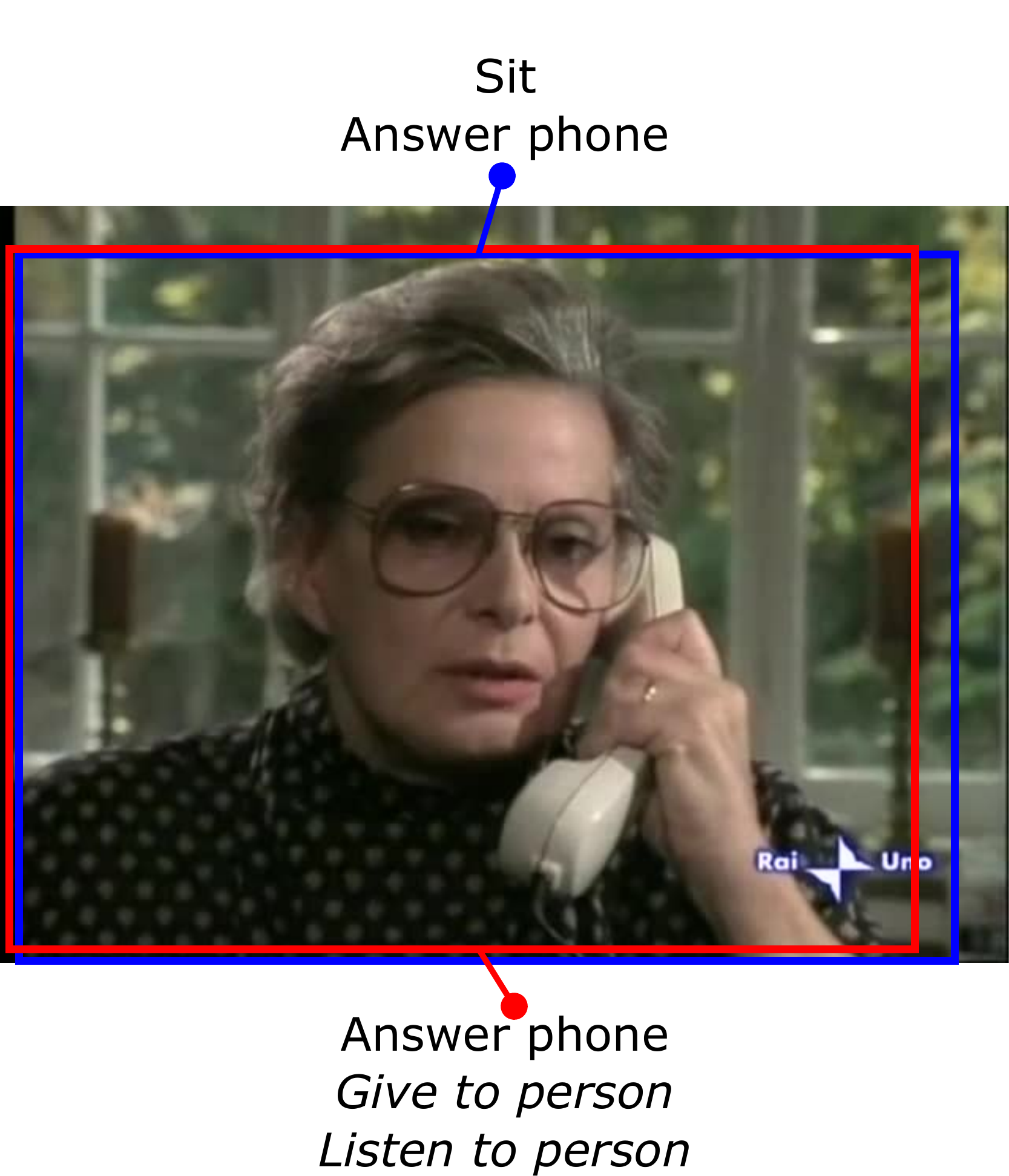}
\caption{}
\end{subfigure}
\hspace{0.75cm}
\begin{subfigure}{0.265\textwidth}
\includegraphics[width=\textwidth]{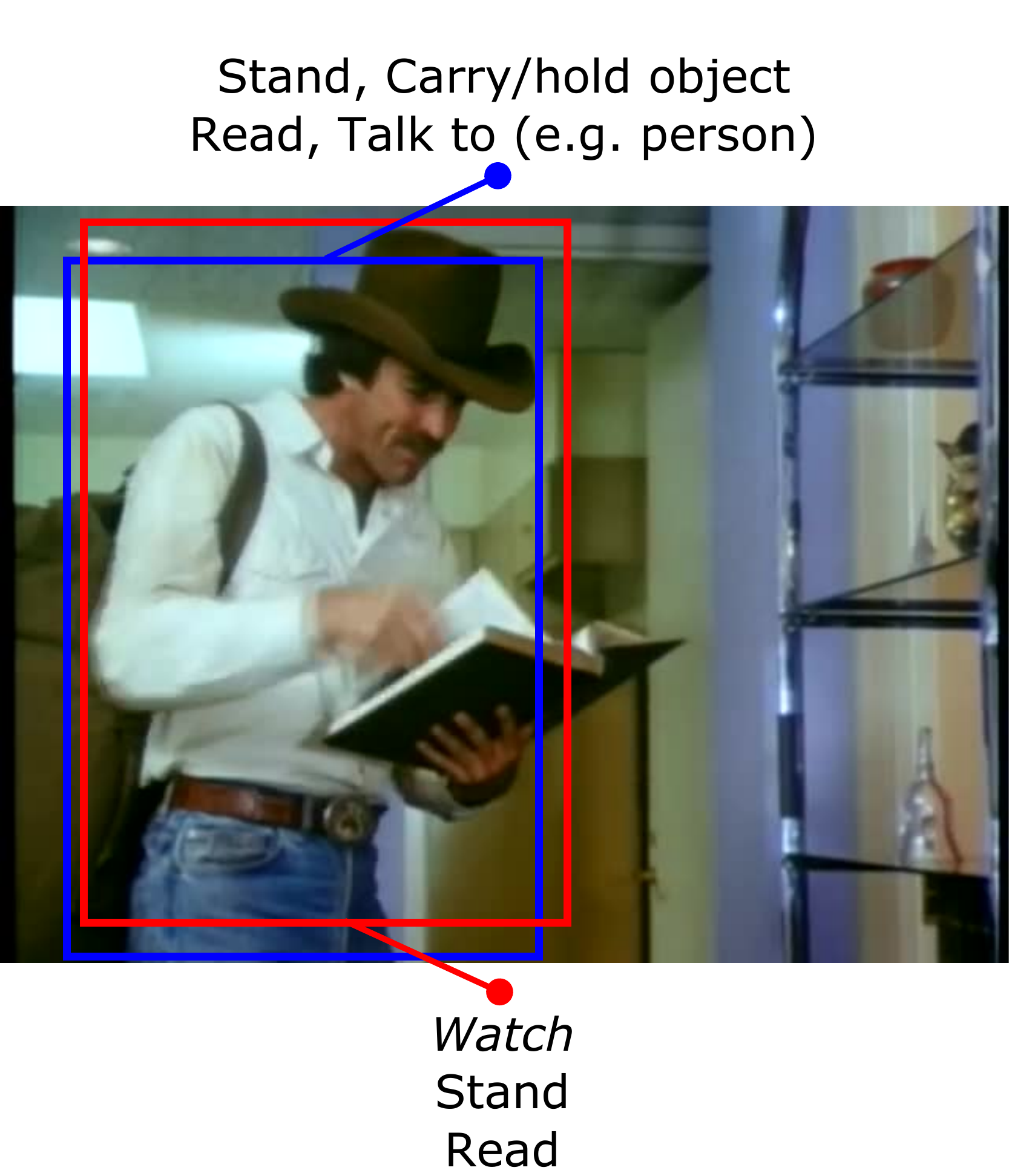}
\caption{}
\end{subfigure}
\hspace{0.75cm}
\begin{subfigure}{0.265\textwidth}
\includegraphics[width=\textwidth]{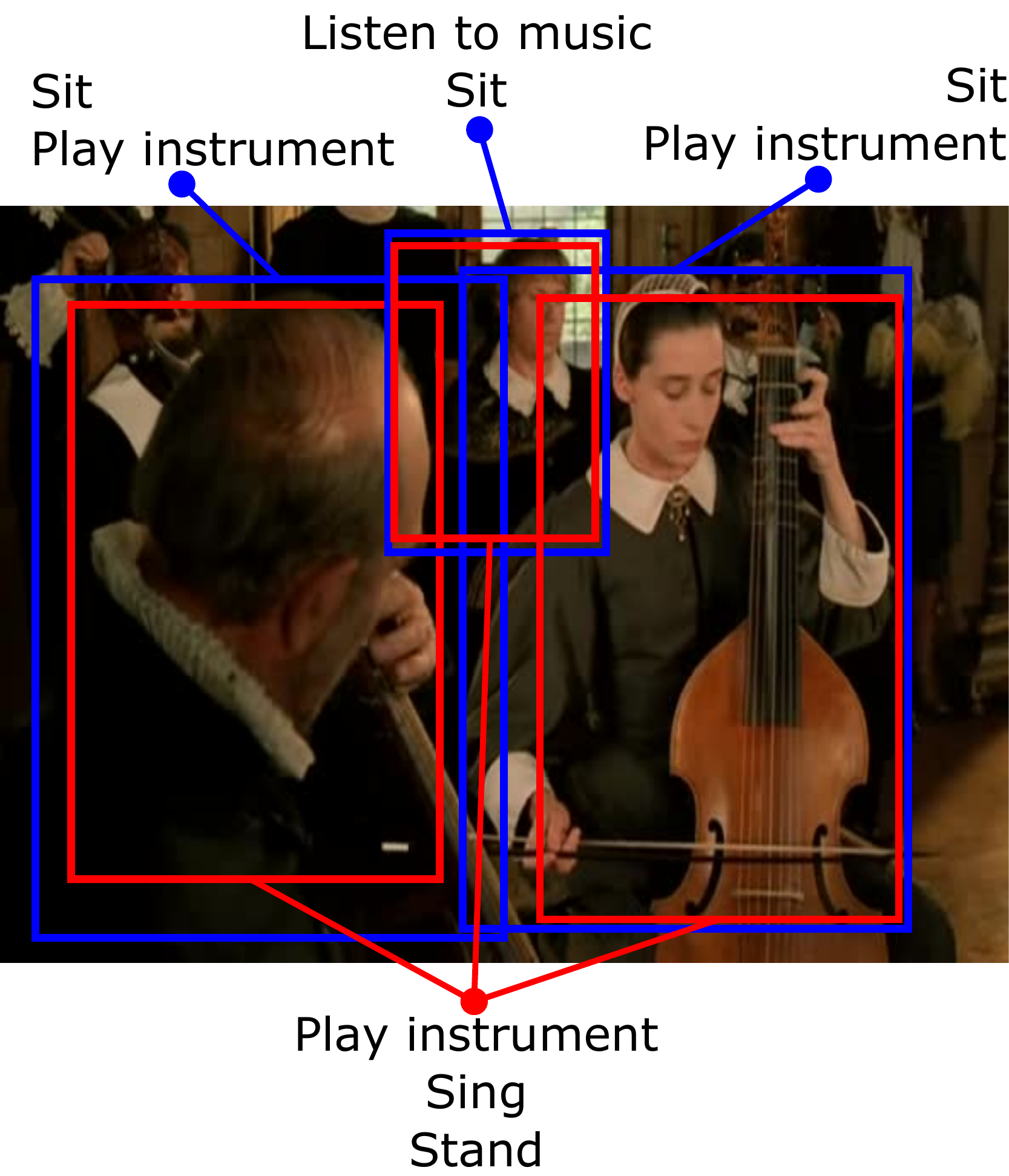}
\caption{}
\end{subfigure}
\caption{\oldreb{Challenges for unseen action localization with object priors in the wild on AVA keyframes~\citep{gu2018ava}. For each keyframe, we show the top three highest scoring actions (below frame) for the detected persons (red boxes), compared to the ground truth actions (above frame and blue boxes). In all three keyframes, at least one ground truth action is in our top actions due to relevant objects, resp. a phone in (a), a book in (b), and an instrument in (c). The keyframes also show open challenges, \eg: it is unknown how many actions are relevant in a frame (a-c), person-centric actions are often missed (talk to in b and sit in c), and fine-grained actions can not be distinguished (listed to music versus playing instrument in c).}}
\label{fig:ava}
\end{figure*}
%\reb{In Figure~\ref{fig:kinetics}, we show unseen action recognition results on Kinetics. Different from, (Hahn et al, 2019), our approach does not require any seen training actions and videos, enabling for the first time a 400-way unseen action recognition. On the full dataset with 400 actions, we obtain an accuracy of 6.41\%, compared to 0.25\% for random performance.}
%
\\\\
\textbf{Unseen action localization.}
In Table~\ref{tab:sota-localization-2}, we show the results for unseen action localization on UCF Sports and J-HMDB. The comparison is made to the only two previous papers with unseen localization results~\citep{jain2015objects2action,mettes2017spatial}. On UCF Sports, we obtain an AUC score of 33.1\%, compared to 7.2\% for~\cite{jain2015objects2action}. We also outperform our previous work~\citep{mettes2017spatial}, using spatial object priors only, by 2\%, reiterating the empirical effect of semantic object priors. We furthermore provide mAP scores on both UCF Sports and J-HMDB. The larger gap in scores compared to the AUC metric on UCF Sports shows that we are now better at ranking correct action localizations at the top of the list for actions. Similarly for J-HMDB, we find consistent improvements across all overlap thresholds, highlighting our effectiveness for unseen action localization.
We conclude that object priors matter for unseen action classification and localization, resulting in state-of-the-art scores on both tasks.
%\reb{Figure~\ref{fig:ava} highlights a number of open challenges for unseen action localization.}

\reb{Next to unseen action localization experiments on UCF Sports and J-HMDB, we also provide, for the first time, unseen localization on AVA. In Figure~\ref{fig:ava-perclass}, we show the frame AP for all 80 actions. We obtain a mean AP of 3.7\%, compared to 0.7\% for random scores with the same detected objects and persons. This result shows that large-scale multi-person action localization without training videos is feasible. Our zero-shot approach can identify contextual actions such as \emph{play musical instrument} and \emph{sail boat}, while it struggles with fine-grained actions that focus on person dynamics instead of object interaction, such as \emph{crawl} and \emph{fall down}.}

\reb{The quantitative results on AVA show that large-scale unseen action localization is feasible, but multiple open challenges remain. In Figure~\ref{fig:ava}, we highlight three open challenges to improve localization performance. Most notably, it is unknown in the zero-shot setting how many actions occur at each timestep, while person-centric actions are often missed due to the lack of informative objects and context. Fine-grained actions (\eg \emph{listen to} versus \emph{playing music}) are also difficult in dense scenes. Addressing these challenges require priors that go beyond objects, including but not limited to action priors and person skeleton priors.}

\section{Conclusions}
\label{sec:conclusions}
This work advocates the importance of using priors obtained from objects to enable unseen action classification and localization. We propose three spatial object priors, allowing for spatio-temporal localization without examples. Additionally, we propose three semantic object priors to deal with semantic ambiguity, object discrimination, and object naming in the semantic matching. Even though no video examples are available during training, the object priors provide strong indications what actions happen where in videos. Due to the generic setup of our priors, we also introduce a new task, action tube retrieval, where users specify object type, spatial relations, and object size to obtain spatio-temporal locations on-the-fly. The use of spatial and semantic object priors results in state-of-the-art scores for unseen action classification and localization.
\oldreb{We conclude that objects make sense for unseen actions when the set of actions is heterogeneous, as is the case in common action datasets. When actions become more fine-grained, \eg throwing versus catching a ball, spatial and semantic priors alone might not be sufficient, urging the need for causal temporal priors about objects and persons. For zero-shot interactions between persons, a fruitful source of priors to explore relate to knowledge about body pose.}
\bibliographystyle{spbasic}      % basic style, author-year citations
\bibliography{egbib}   % name your BibTeX data base

\end{document}